\documentclass[10pt,twocolumn,letterpaper]{article}

\usepackage[pagenumbers]{cvpr} 

\usepackage{graphicx}
\usepackage{amsmath}
\usepackage{amssymb}
\usepackage{booktabs}

\usepackage[pagebackref,breaklinks,colorlinks]{hyperref}

\usepackage{animate}
\makeatletter
\@namedef{ver@everyshi.sty}{}
\makeatother
\usepackage{tikz}
\usepackage{multirow}
\usepackage{bm}
\usepackage{fixmath}
\usepackage[capitalize]{cleveref}
\Crefname{section}{Section}{Sections}
\Crefname{table}{Table}{Tables}
\crefname{table}{Tab.}{Tabs.}
\crefname{section}{Sec.}{Secs.}
\usepackage{xspace}
\makeatletter
\DeclareRobustCommand\onedot{\futurelet\@let@token\@onedot}
\def\@onedot{\ifx\@let@token.\else.\null\fi\xspace}
\def\eg{\emph{e.g}\onedot} 
\def\ie{\emph{i.e}\onedot}

\def\etal{\emph{et al}\onedot}
\makeatother

\usepackage{pifont}
\usepackage{colortbl}
\definecolor{Gray}{gray}{0.9}

\def\cvprPaperID{*****} 
\def\confName{CVPR}
\def\confYear{2023}

\begin{document}

\title{NeRFPlayer: A Streamable Dynamic Scene Representation with \\ Decomposed Neural Radiance Fields}

\author{
Liangchen Song$^{1}$\quad Anpei Chen$^{2,4}$\quad Zhong Li$^{3}$\quad Zhang Chen$^{3}$\quad Lele Chen$^{3}$\\ Junsong Yuan$^{1}$\quad Yi Xu$^{3}$\quad Andreas Geiger$^{4}$\\
{\small $^1$University at Buffalo $\; ^2$ETH Zürich}\\
{\small$\; ^3$OPPO US Research Center, InnoPeak Tech $\; ^4$University of Tübingen} \\
\href{https://lsongx.github.io/projects/nerfplayer.html}{https://bit.ly/nerfplayer}
}

\twocolumn[{%
\renewcommand\twocolumn[1][]{#1}%
\maketitle
\vspace*{-3em}
\begin{center}
    \centering
    \captionsetup{type=figure}
      \resizebox{\textwidth}{!}{
  \begin{animateinline}[autoplay,loop,palindrome,poster=6]{3}
    \centering
    \multiframe{9}{i=0+1}{%
        \begin{tabular}{ccc}
        \includegraphics[height=1.29in]{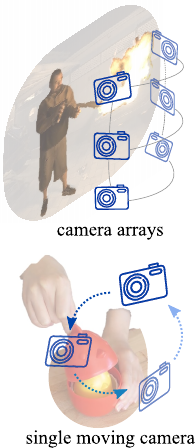} 
        & 
        \includegraphics[height=1.29in]{fig/guigif/flames\i}
        \includegraphics[height=1.29in]{fig/guigif/chicken\i}
        & 
        \begin{tikzpicture}
          \draw (0, 0) node[inner sep=0] {\includegraphics[height=1.29in]{fig/bitrategif/truck\i}};
          \fill [white] (-0.6, -1.58) rectangle (0.6,-1.42);
          \draw (0, -1.5) node {\tiny 59.8 KB/frame};
        \end{tikzpicture}
        \begin{tikzpicture}
          \draw (0, 0) node[inner sep=0] {\includegraphics[height=1.29in]{fig/bitrategif/salmon\i}};
          \fill [white] (-0.6, -1.58) rectangle (0.6,-1.42);
          \draw (0, -1.5) node {\tiny 78.8 KB/frame};
        \end{tikzpicture}
        \\
        {\small (a) Inputs} &
        {\small (b) Real-time rendering} &
        {\small (c) Low bitrate streaming}
        \end{tabular}
  }
  \end{animateinline}}
    \captionof{figure}{(a) Our framework takes as input the RGB images captured from a camera array or a single moving camera. (b) After offline optimization, our framework can render a novel view and perform temporal interpolation interactively. (c) Our framework is highly configurable. Adopting TensoRF-CP \cite{tensorf} voxel representation in our framework results in low bitrate streaming of high-quality rendering.}
    \label{fig:teaser}
\end{center}%
}]

\begin{abstract}
Visually exploring in a real-world 4D spatiotemporal space freely in VR has been a long-term quest. The task is especially appealing when only a few or even single RGB cameras are used for capturing the dynamic scene. To this end, we present an efficient framework capable of fast reconstruction, compact modeling, and streamable rendering. First, we propose to decompose the 4D spatiotemporal space according to temporal characteristics. Points in the 4D space are associated with probabilities of belonging to three categories: static, deforming, and new areas. Each area is represented and regularized by a separate neural field. Second, we propose a hybrid representations based feature streaming scheme for efficiently modeling the neural fields. Our approach, coined NeRFPlayer, is evaluated on dynamic scenes captured by single hand-held cameras and multi-camera arrays, achieving comparable or superior rendering performance in terms of quality and speed comparable to recent state-of-the-art methods, achieving reconstruction in 10 seconds per frame and interactive rendering.
\end{abstract}
\let\thefootnote\relax\footnote{Work done while the first author was an intern at Innopeak Tech.}

\section{Introduction}
Representing scenes as Neural Radiance Fields (NeRF) has brought a series of breakthroughs in 3D reconstruction and analysis \cite{xie2021neural,mildenhall2020nerf}. High-fidelity real-time rendering of real-world scenes now can be obtained after a few seconds of training \cite{yu_and_fridovichkeil2021plenoxels,instantngp}. The rendering system only requires a few real-world RGB images \cite{mildenhall2021rawnerf}, but can well model  scenes as small as a cell \cite{liu2022recovery} and as large as a city \cite{tancik2022blocknerf} or even a black hole \cite{levis2022gravitationally}.

Despite NeRF's success in static scenes, extending it to handle dynamic scenes remains challenging. Introducing an extra time dimension $t$ to NeRF's 5D representation (3D location $x,y,z$ and 2D viewing direction $\theta,\phi$) is non-trivial for the following two reasons. 
First, the supervisory signal for a spatiotemporal point $(x,y,z,t)$ is sparser than a static point $(x,y,z)$. Multi-view images of static scenes are easy to access as we can move the camera around, but an extra view in dynamic scenes requires an extra recording camera, leading to sparse input views. 
Second, the appearance and geometry frequency of the scene are different along the spatial axis and temporal axis. The content usually changes a lot when moving from one location to another location, but the background scene is unlikely to completely change from one timestamp to another. An inappropriate frequency modeling for the time $t$ dimension results in poor temporal interpolation performance.

A lot of progress has been made in addressing the aforementioned two challenges. 
Existing solutions include adopting motion models for matching the points (\eg, \cite{pumarola2020dnerf,nerfies,Tretschk_2021_ICCV,park2021hypernerf,liu2022devrf,fang2022fast}) and leveraging data-driven priors like depth and optical flow (\eg, \cite{wang2021neural,nsff,xian2020space,du2021nerflow}).
Different from existing works, we are motivated by the observation that in dynamic scenes different spatial areas have different temporal characteristics.
We assume that there are three kinds of temporal patterns in a dynamic scene (\cref{fig:points,fig:toy-example}): static, deforming, and new areas. 
We thus propose to decompose the dynamic scene into these categories, which is achieved by a decomposition field that predicts the point-wise probabilities of being static, deforming, and new. 
The decomposition field is self-supervised and regularized by a manually assigned global parsimony regularization (\eg, suppressing the global probabilities of being new).

The proposed decomposition can address both of the aforementioned challenges. First, different temporal regularizations are introduced for each decomposed area, thus alleviating the ambiguity in reconstruction from sparse observations. For instance, the static area decomposition simplifies the dynamic modeling to a static scene modeling problem. The deforming areas enforce the foreground object to be consistent in the dynamic scene. Second, the scene is split into different areas according to their temporal characteristics, thus resulting in consistent frequency in the time dimension in each of the areas.

In response to the discrepancy between spatial and temporal frequency, we further decouple spatial and temporal dimensions based on the recent developed hybrid representations \cite{dvgo,yu_and_fridovichkeil2021plenoxels,instantngp,tensorf}. Hybrid representations maintain a grid of $(x,y,z)$ feature volumes for fast rendering.
Instead of designing a grid of $(x,y,z,t)$ feature volumes, we treat the channels of $(x,y,z)$ feature volumes as temporally dependent. To support streamable dynamic scene representation, we propose a sliding window scheme on the feature channels to introduce $t$ into the representation (\cref{fig:stream}). Sliding window not only supports streaming of the feature volumes, but also implicitly encourages the representation to be compact by leveraging the overlapped channels in adjacent frames.

For validation, we conduct experiments on datasets captured under both single-camera and multi-camera settings. Our extensive ablation studies validate our proposed method in three aspects: 1) the necessity of modeling all of the three areas on single-camera datasets, 2) the necessity of decomposing static areas on multi-camera datasets, and 3) the necessity of deforming decomposition on inputs with large frame-wise motion even for multi-camera datasets.
To sum up, our contributions are as follows:
\begin{itemize}
  \item We propose to decompose the dynamic scene according to their temporal characteristics. The decomposition is achieved by a decomposition field that takes as input each $(x,y,z,t)$ point and outputs probabilities belonging to three categories: static, deforming, and new. 
  \item We design a self-supervised scheme for optimizing the decomposition field and regularizing the decomposition field with a global parsimony loss.
  \item We design a sliding window scheme on recently developed hybrid representations for efficiently modeling spatiotemporal fields. 
  \item We present extensive experiments and interactive rendering demos on both single-camera and multi-camera datasets. Our ablation studies validate the implied regularizations behind the proposed three temporal patterns.
\end{itemize}

\section{Related Work}
\subsection{Neural Fields}
Neural fields are neural networks that take in the coordinates and output the properties of that point \cite{xie2021neural}. 3D representations based on neural fields have made tremendous advancements in recent years. The pioneering work Occupancy Networks \cite{mescheder2019occupancy} represents the geometry of 3D objects with a continuous decision boundary modeled by a neural network. Occupancy Networks are further improved to model dynamic objects \cite{niemeyer2019occupancy}. Concurrently, DeepSDF \cite{park2019deepsdf} represents the geometry with signed distance function with a network. Chibane \etal \cite{chibane2020neural} predicts the unsigned distance field for 3D shapes from point clouds. NeRF \cite{mildenhall2020nerf}, a milestone work, proposes to represent the scene with a 5D function modeled by MLP. NeRF significantly improves the performance of novel view synthesis (\ie, image-based rendering). The scene representation in NeRF inspired a number of works focusing on 3D modeling, such as human face and body capture \cite{hong2021headnerf,Noguchi_2021_ICCV,peng2021neural,Peng_2021_ICCV,su2021anerf,liu2021neural}, relighting \cite{boss2021nerd,srinivasan2021nerv,boss2021neuralpil} and 3D content generation \cite{Trevithick_2021_ICCV,schwarz2020graf,chan2021pi,gu2021stylenerf,kosiorek2021nerf,chan2022efficient,dreamfields}.
\paragraph{Hybrid Representation}
Scenes are implicitly represented by MLPs in vanilla NeRF and forwarding with the MLPs is time-consuming. Some methods like DONeRF \cite{neff2021donerf} accelerate the sampling step \cite{autoint,fang2021neusample,Piala2021TermiNeRFRT,Kurz2022AdaNeRFAS}. HyperReel \cite{attal2023hyperreel} and ENeRF \cite{lin2022efficient} have adapted this idea in dynamic scenes. Another set of insightful methods \cite{nsvf,wizadwongsa2021nex,PlenOctrees,Reiser_2021_ICCV,Hedman_2021_ICCV,Garbin_2021_ICCV,wu2021diver} are designed by adopting explicit data structures to efficiently query from the fields. Further, hybrid representations are developed by leveraging both explicit and implicit representations to improve the differentiability of the framework. DVGO \cite{dvgo} uses two feature voxels to represent occupancy and appearance. The feature vectors queried from the voxels are decoded by small MLPs. Plenoxels \cite{yu_and_fridovichkeil2021plenoxels} prune empty spaces and save the sphere harmonic coefficients. InstantNGP \cite{instantngp} proposes a hash encoding of the saved feature grids and solves hashing collision by multi-scale encoding and small MLP decoding. TensoRF \cite{tensorf} leverages tensor decomposition to reduce the model size of the voxels. Hybrid representations are further leveraged for efficient dynamic scene modeling. Recent concurrent works \cite{fang2022fast,liu2022devrf,gan2022v4d} propose to model canonical spaces with voxels and motion with deformation fields. Li \etal \cite{streamrf} propose to stream the difference of voxels in dynamic scenes. Different from the above methods, our method decomposes the scene into different areas and models them separately.
A straightforward InstantNGP based dynamic representation is adding extra input dimension of time, but such a baseline requires the full representation of a dynamic sequence to be completely loaded into the GPU memory before rendering. For TensoRF based dynamic modeling, D-TensoRF \cite{Jang2022DTensoRFTR} uses a 5D tensor to represent a 4D spatiotemporal grid. HexPlane \cite{HexPlane} and K-Planes \cite{kplanes} propose to decompose the dynamic scene to a set of planes. Our method can be widely applicable, as long as the representation adopts feature vectors for modeling points in the space.
\paragraph{Scene Decomposition}
Neural fields have been adopted for decomposing scenes. Yang \etal \cite{yang2021learning} and Zhang \etal \cite{ZhangLYZZWZXY21} decompose the scene by objects for editing. DeRF \cite{rebain2021derf} spatially decomposes the scene and uses small networks for each area for efficiency. Kobayashi \etal \cite{kobayashi2022distilledfeaturefields} and Tschernezki \etal \cite{tschernezki22neural} semantically decompose the scene with pre-trained models. Ost \etal \cite{ost2021neural} decompose scenes into semantic scene graphs. Objects are decomposed by motion in NeuralDiff \cite{tschernezki2021neuraldiff} and STaR \cite{yuan2021star}. More recently, a decomposition between static and dynamic areas is studied in D$^{2}$NeRF \cite{wu2022d} and Sharma \etal \cite{sharma2022seeing}. Our decomposition is different from existing works since we decompose areas according to the temporal changing patterns.
\subsection{4D Modeling of Dynamic Scenes}
Free viewpoint rendering from video captures has been widely studied over the decades. The idea of viewing an event from multiple perspectives dates back to  Multiple Perspective Interactive Video \cite{jain1995multiple}, in which 3D environments are generated with dynamic motion models. Virtualized Reality \cite{kanade1997virtualized} design 3D dome and recovered 3D structure based on multi-camera stereo methods. Inspired by image-based rendering \cite{LevoyH96,GortlerGSC96}, some video-based rendering methods are developed \cite{schirmacher2001fly,yang2002real,carranza2003free}, which requires dense capturing of the scene. Zitnick \etal \cite{zitnick2004high} propose a layered depth image representation for the high-quality video-based rendering of dynamic scenes. More recently, a milestone work developed by Collet \etal \cite{collet2015high} utilizes tracked textured meshes for free-viewpoint video streaming. With RGB, infrared (IR), and silhouette information as the input, their system can output accurate geometric, detailed texture, and efficient streaming. Another impressive system developed by Broxton \etal \cite{broxton2020immersive} proposes multi-sphere image based Layered Meshes. The capturing setting is a low-cost hemispherical array with 46 synchronized cameras and Layered Meshes are validated to be efficient and can well-handle non-Lambertian surfaces with view-dependent or semi-transparent effects. Bansal \etal \cite{bansal20204d} use convolutional neural nets to compose static and dynamic parts of the event and then adopt U-Net to render images from intermediate results composited from depth-based re-projected images. Neural Volumes (NV) \cite{lombardi2019neural} leverages differentiable volume rendering for optimizing a 3D volume representation, which can be transformed from 2D input RGB images using an encoder-decoder network. NV is further strengthened in \cite{LombardiSSZSS21} with volumetric primitives. X-Fields \cite{bemana2020x} consider input images from different view, time or illumination conditions in structured captures. DyNeRF \cite{Li_2022_CVPR} assign observation frames with a set of compact latent codes and then use time-conditioned neural radiance fields to represent dynamic scenes. Fourier PlenOctrees \cite{wang2022fourier} extend the real-time rendering framework PlenOctrees \cite{PlenOctrees} to dynamic scenes. DeVRF \cite{liu2022devrf} proposes a voxel-based representation that first reconstructs a canonical object from multi-view dense supervisions and then reconstructs deformation from few-view observations. 

Another thread of research aims at modeling dynamic scenes without the requirements of multiple synchronized cameras. Multi-view information is collected by moving the camera around in the dynamic space. The setting of single-camera input is much more challenging than the multi-camera setting mentioned above. Data-driven solutions like video depth estimation \cite{luo2020consistent,kopf2021robust} are developed. Based on the priors and motivated by the success of NeRF, motions are modeled by neural fields. Some methods first define a canonical space that is modeled by a NeRF, then align the following frames from the canonical space. Representative methods include D-NeRF \cite{pumarola2020dnerf}, Nerfies \cite{nerfies} and NR-NeRF \cite{Tretschk_2021_ICCV}. The trajectory of points is modeled by a neural field in DCT-NeRF \cite{wang2021neural}. Directly modeling the 4D field by introducing an extra time dimension into the original radiance field is adopted in NSFF \cite{nsff}, VideoNeRF \cite{xian2020space}, and NeRFlow \cite{du2021nerflow}. HyperNeRF \cite{park2021hypernerf} points out the issue of motion inconsistency in topologically varying scenes and proposes a hyperspace representation, which is inspired by the level-set methods, for optimizing motion in a more smooth solution space. Gao \etal \cite{gao2022dynamic} demonstrate the discrepancy between the casual monocular video and the above existing monocular testing videos.

The above methods are able to generate impressive results under various settings. However, rendering with both single- and multi-camera inputs can be further studied, such as the effectiveness of motion modeling with multi-camera inputs. Moreover, a tradeoff still exists among model size, training and rendering speed, and rendering quality. Our method studies both single- and multi-camera inputs and focuses on efficient and high-quality free-viewpoint video rendering. 

\begin{figure}[t]
  \centering
  \includegraphics[width=0.9\columnwidth]{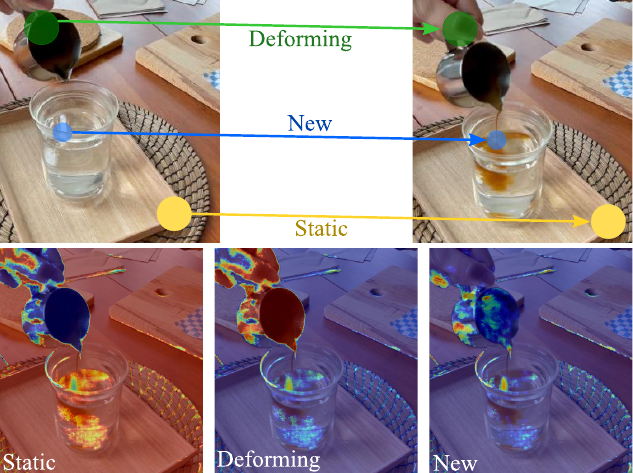}
  \caption{First row: We categorize the areas in a dynamic scene into three groups: deforming, new and static areas. Second row: Visualization of the self-supervised decomposition obtained from our framework. Red and blue areas indicate estimated high and low probabilities of a category.}
  \label{fig:points}
\end{figure}

\begin{figure}[t]
  \centering
  \includegraphics[width=\columnwidth]{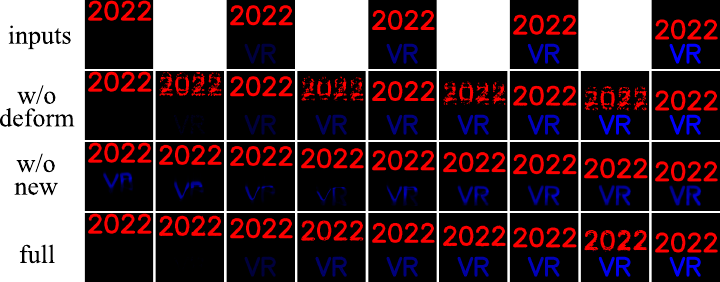}
  \caption{A toy example of 2D dynamic sequence interpolation. The first row shows the 2D input sequence with missing frames. Without modeling deformation $d(\cdot)$, the second row fails to interpolate the rigid motion of `2022'. Without modeling newness $n(\cdot)$, the third row fails to interpolate the gradually appearing effect. Full decomposition handles both phenomena well.}
  \label{fig:toy-example}
\end{figure}

\begin{figure}[t]
  \centering
  \includegraphics[width=0.8\columnwidth]{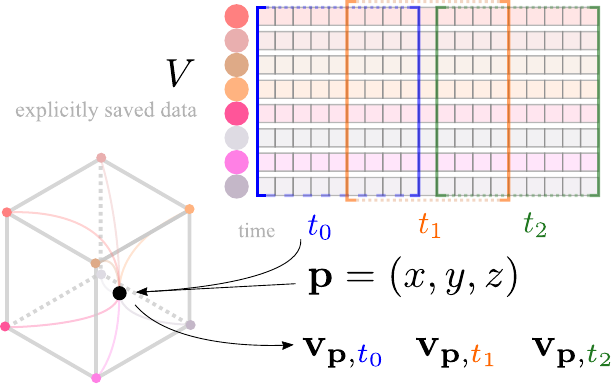}
  \caption{The proposed streamable hybrid representation. A time-dependent sliding window is adopted for streaming the feature channels.}
  \label{fig:stream}
\end{figure}

\begin{figure*}
  \centering
  \includegraphics[width=0.9\textwidth]{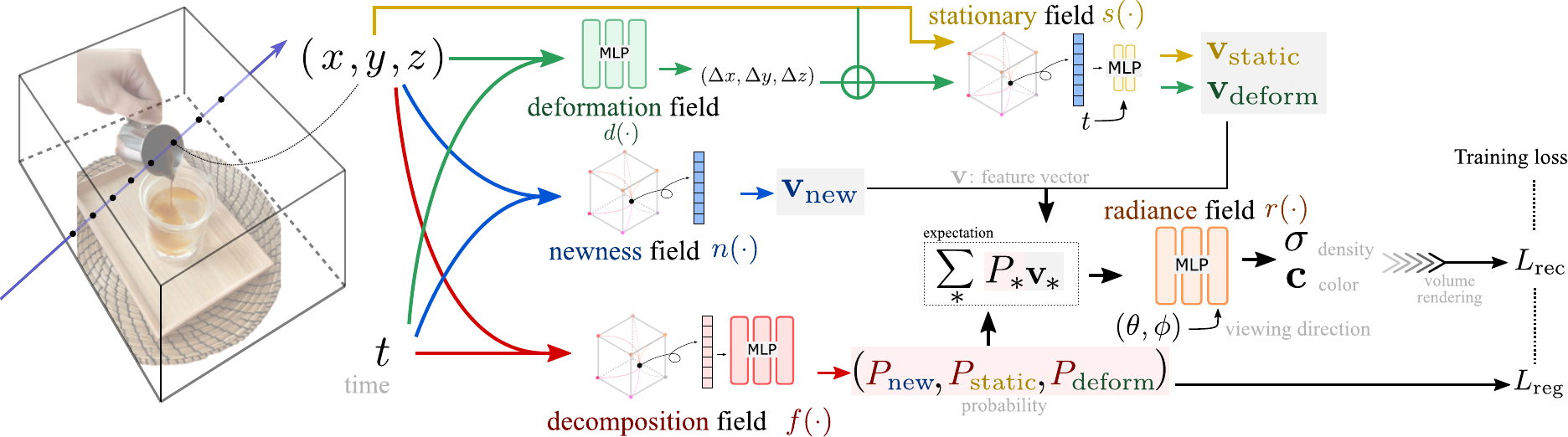}
  \caption{An overview of our framework. The newness field and decomposition field are implemented with the channel streaming technique proposed in \cref{fig:stream}. A small MLP is adopted in the decomposition field for predicting the probabilities. The stationary field consists of a static feature volume for modeling time-invariant areas and a tiny MLP with time $t$ input for modeling low-frequency time-varying appearance. The deformation field and radiance field are two small MLPs.}
  \label{fig:framework}
\end{figure*}

\section{Preliminaries}
Our method leverages the rendering scheme proposed by NeRF \cite{mildenhall2020nerf} and hybrid representation for static scenes \cite{nsvf,takikawa2021neural,dvgo,wu2021diver,yu_and_fridovichkeil2021plenoxels,zhang2022nerfusion,instantngp,takikawa2022variable,tensorf}. We first briefly review the rendering framework in NeRF, then we introduce the recently developed hybrid representation for efficient neural fields.

For each point $\mathbold{p}=(x,y,z)$ in NeRF, we denote its volume density as ${\sigma}(\mathbold{p})$ and its color as $\mathbold{c}(\mathbold{p},\mathbold{d})$, where $\mathbold{d}=(\theta,\phi)$ is the viewing direction.
The pixel color $\mathbold{C}$ of a camera ray $\mathbold{r}$ is computed by accumulating a set of samples on the ray with volume rendering. Let the optical origin and direction of the camera be $\mathbold{o}$ and $\mathbold{d}$, then a set of points are sampled by $\mathbold{p}_i=\mathbold{o}+i\mathbold{d}$ and the expected color $\mathbold{C}(\mathbold{r})$ is computed by
\begin{equation}\label{eq:rendering}
\mathbold{C}(\mathbold{r})=\int_{i_n}^{i_f} e^{-\int_{i_n}^i{\sigma}(\mathbold{p}_j) \mathop{dj}}{\sigma}\big(\mathbold{p}_i\big)\mathbold{c}\big(\mathbold{p}_i,\mathbold{d}\big)\mathop{di},
\end{equation}
where $i_n, i_f$ are near and far bounds. Numerical approximation by summing up a set of sample points on the ray is used for computing the integration in \cref{eq:rendering}. In vanilla NeRF, the radiance field is implicitly represented by an MLP that takes in the point $\mathbold{p}$ as input and outputs its density and color. 
The MLP is then trained with a reconstruction loss between the reconstructed color and ground-truth color $\mathbold{C}_{\mathrm{gt}}(\mathbold{r})$, \ie, 
\begin{equation}\label{eq:recloss}
L_{\mathrm{rec}}=\sum_{\mathbold{r}\in\mathcal{R}}\|\mathbold{C}(\mathbold{r})-\mathbold{C}_{\mathrm{gt}}(\mathbold{r})\|_2^2,
\end{equation}
where $\mathcal{R}$ is a batch of ray samples.


The implicit representation in NeRF is highly compact but computationally expensive, resulting in slow training and rendering speed. Hybrid representations, in which both explicit and implicit representations can be adopted, are developed for efficiently reconstructing and rendering with a radiance field. Though these methods have their unique standouts, all these hybrid representations follow a common framework. 
First, we have some explicitly stored features ${V}$, which can be in the form of a voxel grid \cite{nsvf,takikawa2021neural,dvgo,wu2021diver,yu_and_fridovichkeil2021plenoxels,zhang2022nerfusion}, a hash table \cite{instantngp} or a set of basis vectors/matrices \cite{tensorf}. 
For any point $\mathbold{p}$ in the 3D space, a feature vector $\mathbold{v}_{\mathbold{p}}={V}(\mathbold{p})$ can be efficiently obtained with cheap operations (\eg, tri-linear interpolation for a voxel). Next, a decoder $D$ is adopted to get properties like the density $\sigma$ and color $\mathbold{c}$ of the point from $\mathbold{v}_{\mathbold{p}}$. The decoder $D$ can be an MLP \cite{dvgo,instantngp,tensorf} or spherical harmonics \cite{yu_and_fridovichkeil2021plenoxels}.

\section{Our Method}
Our method is built on the assumption that different areas in a dynamic scene can have different temporal changing patterns.
Modeling different areas with different temporal regularizations not only helps keep temporal consistency but also saves computation. For example, some objects in the background may have a static geometry in the dynamic sequence, which allows us to reduce the capacity and complexity of their representation.
We begin our method with a decomposed spatiotemporal representation which aims to first categorize and then model different dynamic areas using different representations based on their categories.

\subsection{Decomposed Spatiotemporal Representation}\label{sec:decomp}
As illustrated in \cref{fig:points}, we assume three kinds of areas in a dynamic scene and model these areas with separate fields: 
\begin{itemize}
  \item \textbf{Static} areas have a constant geometry and location in the dynamic scene, such as the table. Besides, we assume the appearance of the static areas will not change frequently over time, \ie, is temporally low-frequency. This is based on the observation that the appearance change is mainly caused by lighting conditions and the albedo is time-invariant. Hence a stationary field $s(\cdot)$ is used for representing static points.
  \item \textbf{Deforming} areas model objects with deforming surfaces, such as the hand and the cup in \cref{fig:points}. Deforming areas may comprise rigid or non-rigid motion, but they are always presented in the sequence of interest. 
  Deforming points are represented by a deformation field $d(\cdot): (\mathbold{p},t)\mapsto(\Delta \mathbold{p})$. Then the deformed point $\mathbold{p}+\Delta \mathbold{p}$ is sent as the query point into a predefined canonical space (\eg, the static field $s$). 
  \item \textbf{New} areas model new content emerging at some point in the sequence, such as the new fluid after pouring espresso into water. A newness feature field $n(\cdot)$ with inputs $(\mathbold{p},t)$ is adopted for representing new areas. 
\end{itemize}
To decompose the scene, we design a decomposition field $f(\cdot): (\mathbold{p},t)\mapsto(P_{\mathrm{static}},P_{\mathrm{deform}},P_{\mathrm{new}})$, where $P_{\mathrm{static}},P_{\mathrm{deform}},P_{\mathrm{new}}$ denotes the probability of being static, deforming and new. Next, we consider the output of the fields mentioned above ($s,n$) to be feature vectors rather than properties like the density of the point and denote the output feature vector as $\mathbold{v}_\mathrm{static},\mathbold{v}_\mathrm{deform},\mathbold{v}_\mathrm{new}$, respectively.
Finally, given a query point $\mathbold{p}$, we first collect the outputs from the above fields and then compute the expected feature vector $\mathbold{v}$ of this point by $\mathbold{v}=\sum_\ast P_\ast\mathbold{v}_\ast$, where $\ast\in\{\mathrm{static,deform,new}\}$. Then $\mathbold{v}$ is sent to a lightweight view-conditioned network for density and color prediction.

In \cref{fig:toy-example}, we demonstrate our approach using a simple 2D toy example. The task studied in the figure is a temporal interpolation from the given 2D images. The fields mentioned above take $(x,y)$ locations as the input. The string `2022' undergoes rigid motion while the string `VR' gradually appears. The different interpolation performance demonstrates the necessity of modeling dynamic scenes with both deforming and new fields. Note that we do not manually annotate the probability when performing decomposition. Instead, the decomposition field $f$ is only supervised by the reconstruction loss and generic parsimony priors which penalize objects being modeled as new.
More details about training will be introduced in \cref{sec:optim}.

Hybrid representations, which enable fast training and real-time rendering, are adopted for implementing the above neural fields.
However, most of existing static scene targeted hybrid representations implement the mapping $\mathbold{p}\mapsto(\sigma,\mathbold{c})$. Adapting to inputs with an extra dimension time $t$ (\ie, dynamic scenes) is not straightforward, since modeling 4D inputs with the explicit representation $V$ may significantly increase the model size. A streamable hybrid representation for efficient spatiotemporal mapping is introduced in the next section.

\subsection{Streamable Hybrid Representation}\label{sec:stream}
We observe that the explicit representation $V$ commonly consists of array entries with a predefined feature dimension. For example, each entry in the hash table in InstantNGP \cite{instantngp} and each basis vector/matrix in TensoRF \cite{tensorf} both have a fixed feature dimension. 
Thus, we propose to stream the \textit{feature channels} so that $V$ can be a mapping from a spatiotemporal point $(\mathbold{p},t)$ to the fixed-length feature vector $\mathbold{v}_{\mathbold{p},t}$. 

We propose to select feature channels with a sliding window along with the time dimension $t$, as demonstrated in \cref{fig:stream}.
Assume that for each frame the feature vector $\mathbold{v}_{\mathbold{p},t}$ is of dimension $F$ and $k$ channels are newly needed for a new frame, then for a $T$ frame sequence the array entry $v$ in $V$ is of dimension $F+k(T-1)$. For a single frame $t$, the channels $[kt,kt+F]$ in $V$ will be used for computing $\mathbold{v}_{\mathbold{p},t}$, such as trilinear interpolation in InstantNGP or tensor multiplication in TensoRF. 

To ensure $\mathbold{v}_{\mathbold{p},t}$ smoothly translates along with $t$, a rearrangement of feature channels is conducted to match the shared channels. For example, let $t=0,k=2,$ and $F=4$, then channels $[0,1,2,3]$ of $V$ are used for $\mathbold{v}_{\mathbold{p},0}$. Next, we use channels $[4,5,2,3]$ for $t=1$ and channels $[4,5,6,7]$ for $t=2$. The principle behind the rearrangement is that shared feature channels are always aligned to be with the same index in the vector. Otherwise, a smooth translation between frames is not guaranteed.

The streaming channels readily enable us to temporally interpolate a frame $t$ between two observed frames $t_s$ and $t_{s+1}$ by linearly interpolating the feature vectors: $\mathbold{v}_{\mathbold{p},t}=\frac{t-t_{s}}{t_{s+1}-t_s}\mathbold{v}_{\mathbold{p},t_{s+1}}+\frac{t_{s+1}-t}{t_{s+1}-t_s}\mathbold{v}_{\mathbold{p},t_{s}}$. 
Note that our proposed method can be applied to any hybrid representation $V$ that contains entries of feature vectors. The implementation of $V$ employed will be referred to as \textit{backbone} in the following text.
The sliding window scheme brings two benefits: First, overlapping feature channels are forced to be shared in adjacent frames, thus reducing the model size; Second, after rendering one frame, only new feature channels need to be loaded when moving to the subsequent frames, thus being streaming friendly.

\subsection{Overall Framework.}\label{sec:framework}
Now we introduce the details of implementing the decomposed spatiotemporal representation (\cref{sec:decomp}) with the streamable hybrid representation (\cref{sec:stream}). An illustration is presented in \cref{fig:framework}. The decomposition field $f$ consists of explicitly cached features (denoted by $V_f$) and a small MLP decoder $D_f$. The deformation field $d$ is an MLP since the deformation is sparse and of low-frequency, where a small MLP is enough. The stationary field consists of explicitly cached features (denoted by $V_s$) and a tiny MLP decoder. Time $t$ and feature obtained from $V_s$ will be the input to the tiny MLP. The reason for using a tiny MLP is for modeling time-dependent appearance changes caused by time-varying illumination, which is assumed to be of low-frequency. The newness field $n$ is explicitly saved features $V_n$. Note that in the above explicit representations, both $V_f$ and $V_n$ take in a 4D input $(\mathbold{p},t)$, hence streaming channels are used here.
The final expected feature vector $\mathbold{v}$ is then decoded by a radiance field $r$. Viewing direction $(\theta,\phi)$ is also sent to $r$ as in NeRF.

\subsection{Optimization}\label{sec:optim}
\paragraph{Training.} Our training process follows the practice of NeRF. A batch of camera rays $\mathcal{R}$ is first randomly sampled from the observed data and then points on those rays, denoted by $\mathcal{R}_{\mathbold{p}}$, are sampled for training. 

In practical reconstruction and rendering tasks, a precise supervisory signal to the decomposition field is inaccessible. Instead, we supervise the output probabilities with a global parsimony regularization. 
Therefore, besides the reconstruction loss defined in \cref{eq:recloss}, a regularization loss $L_{\mathrm{reg}}$ is introduced in our method. We use the average probability of all points in the batch for this loss, denoted as $\overline{P_\ast}=\frac{1}{|\mathcal{R}_{\mathbold{p}}|}\sum_{\mathbold{p}\in\mathcal{R}_{\mathbold{p}}}P_{\ast}(\mathbold{p})$, where $|\mathcal{R}_{\mathbold{p}}|$ is the number of points. In our implementation, assuming the existence of static background, we propose to minimize the probability of not being a static point, thus the regularization loss is chosen as
\begin{equation}\label{eq:lossreg}
L_{\mathrm{reg}}=\alpha\overline{P_\mathrm{deform}}+\overline{P_\mathrm{new}},
\end{equation}
where $\alpha$ is a tunable parameter for weighting the ratio of being deforming and new.
Minimizing the probability of being new points in \cref{eq:lossreg} relies on the assumption that most of the points in the dynamic scene are either static or deforming. 
Overall, our training loss is
\begin{equation}
L=L_{\mathrm{rec}}+\lambda L_{\mathrm{reg}},
\end{equation}
where $\lambda$ is a balancing hyper-parameter.

\paragraph{Rendering.} When rendering an image with a given camera pose, we first forward the sampled points using the decomposition field. After knowing the probabilities, we can skip the forwarding process of some fields for efficiency. With a predefined threshold $\tau$, if $P_\mathrm{*}<\tau$ then we directly set $\mathbold{v}_\mathrm{*}=\mathbold{0}$ and skip the field. We set $\tau$ as 0.001 in our implementation.

\section{Experiments}
We first quantitatively and qualitatively compare our method with prior works, then extensive ablation studies are presented to validate our proposed components.
We urge the reader to watch our video to better appreciate the efficiency and rendering quality of our system.

\begin{figure}
\centering

\begin{tabular}{lr}
\hspace{-1em}\begin{tabular}{c}\includegraphics[width=0.23\textwidth]{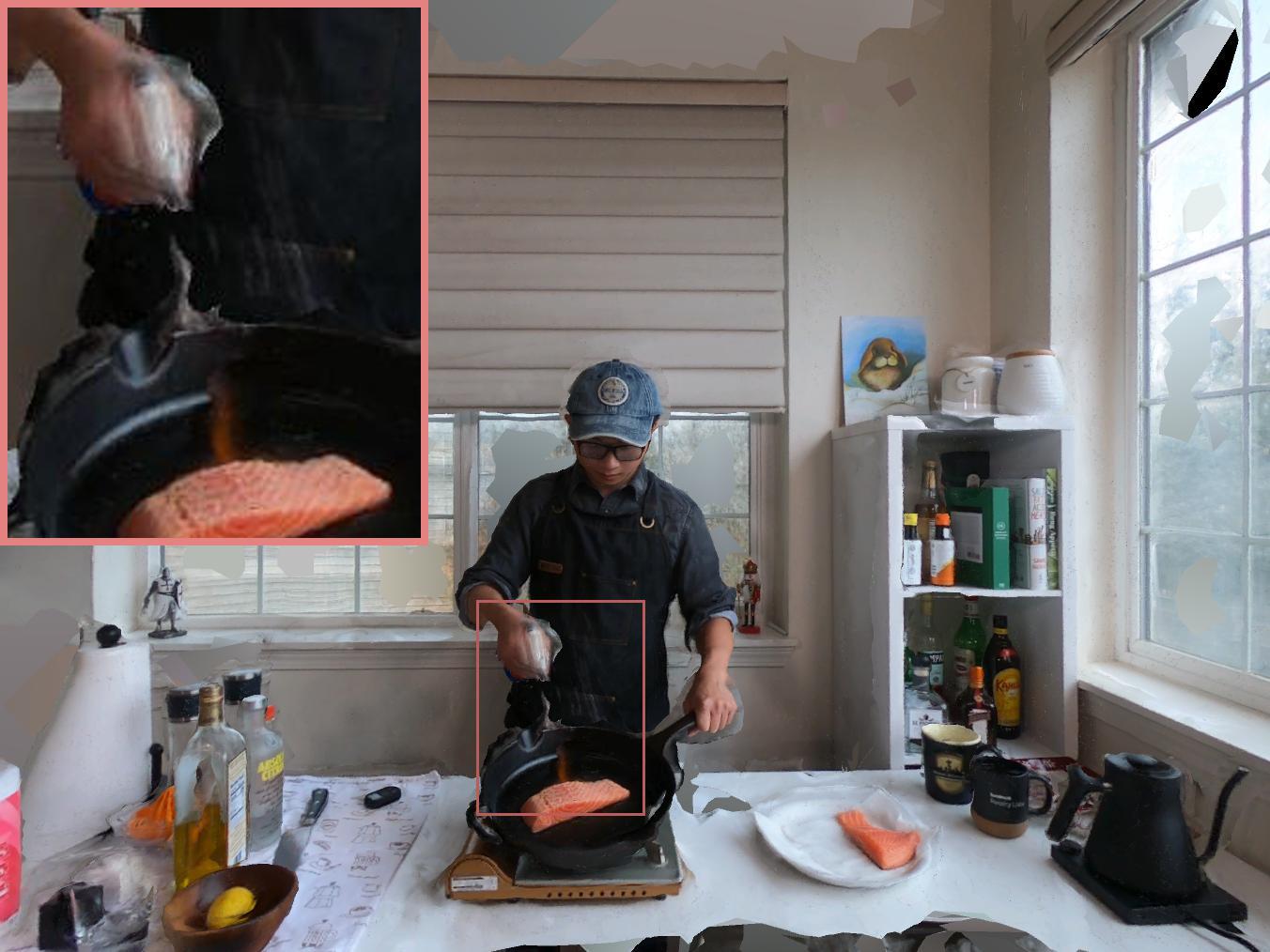}\\Multi-View Stereo\end{tabular} \hspace{-1em} &
\hspace{-1em}\begin{tabular}{c}\includegraphics[width=0.23\textwidth]{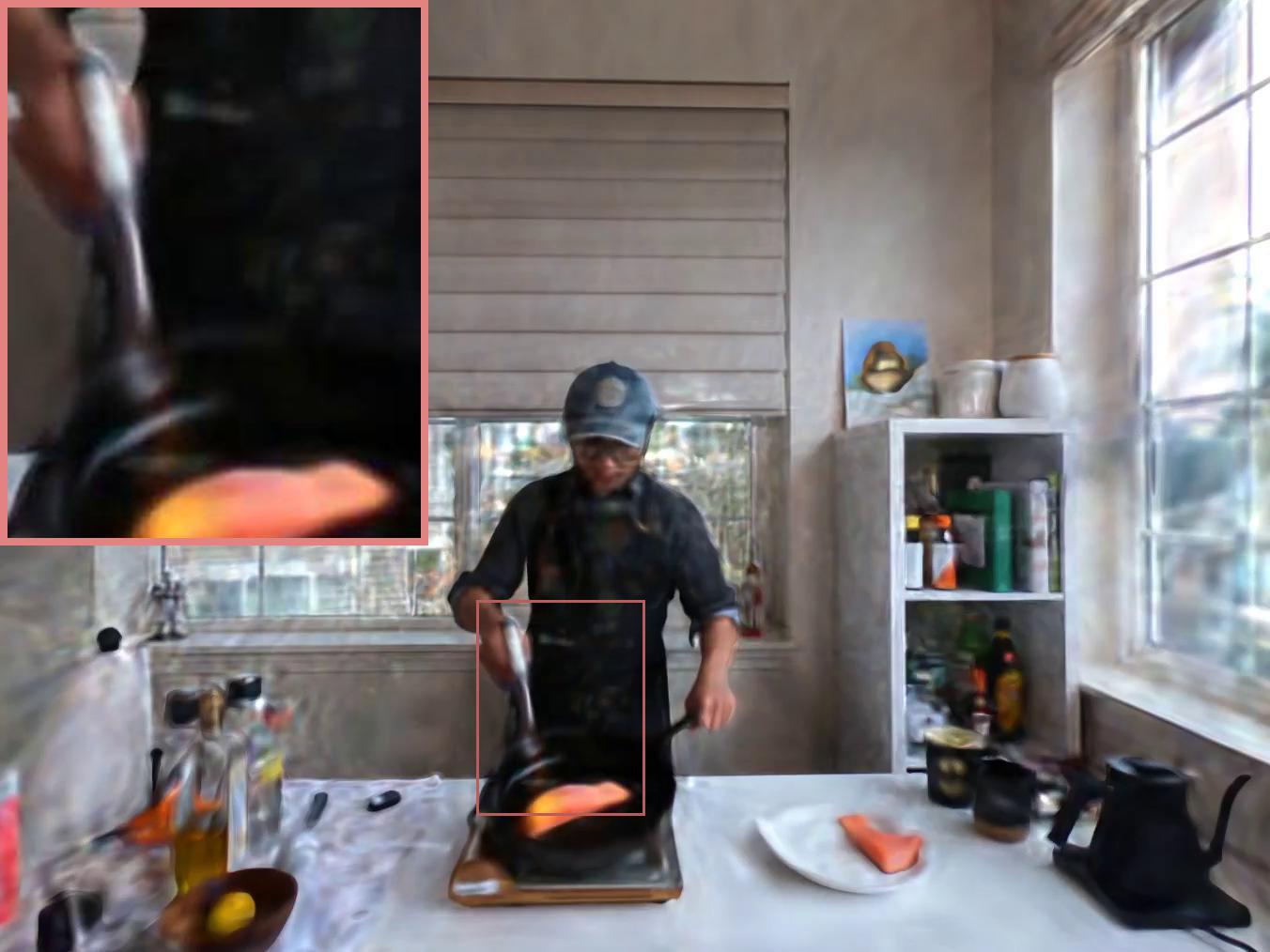}\\NV\cite{lombardi2019neural}\end{tabular} \\
\hspace{-1em}\begin{tabular}{c}\includegraphics[width=0.23\textwidth]{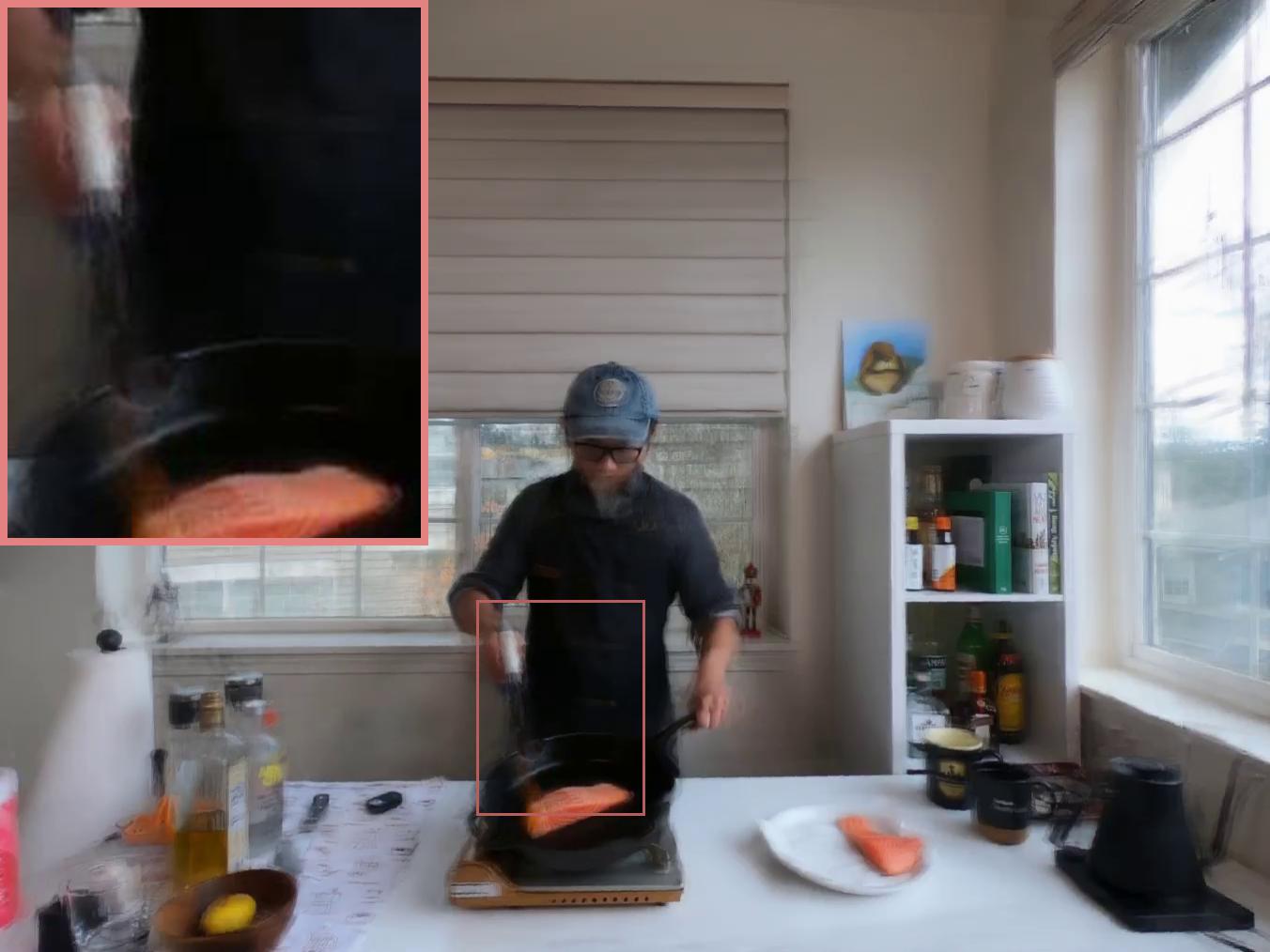}\\LLFF\cite{llff}\end{tabular} \hspace{-1em} &
\hspace{-1em}\begin{tabular}{c}\includegraphics[width=0.23\textwidth]{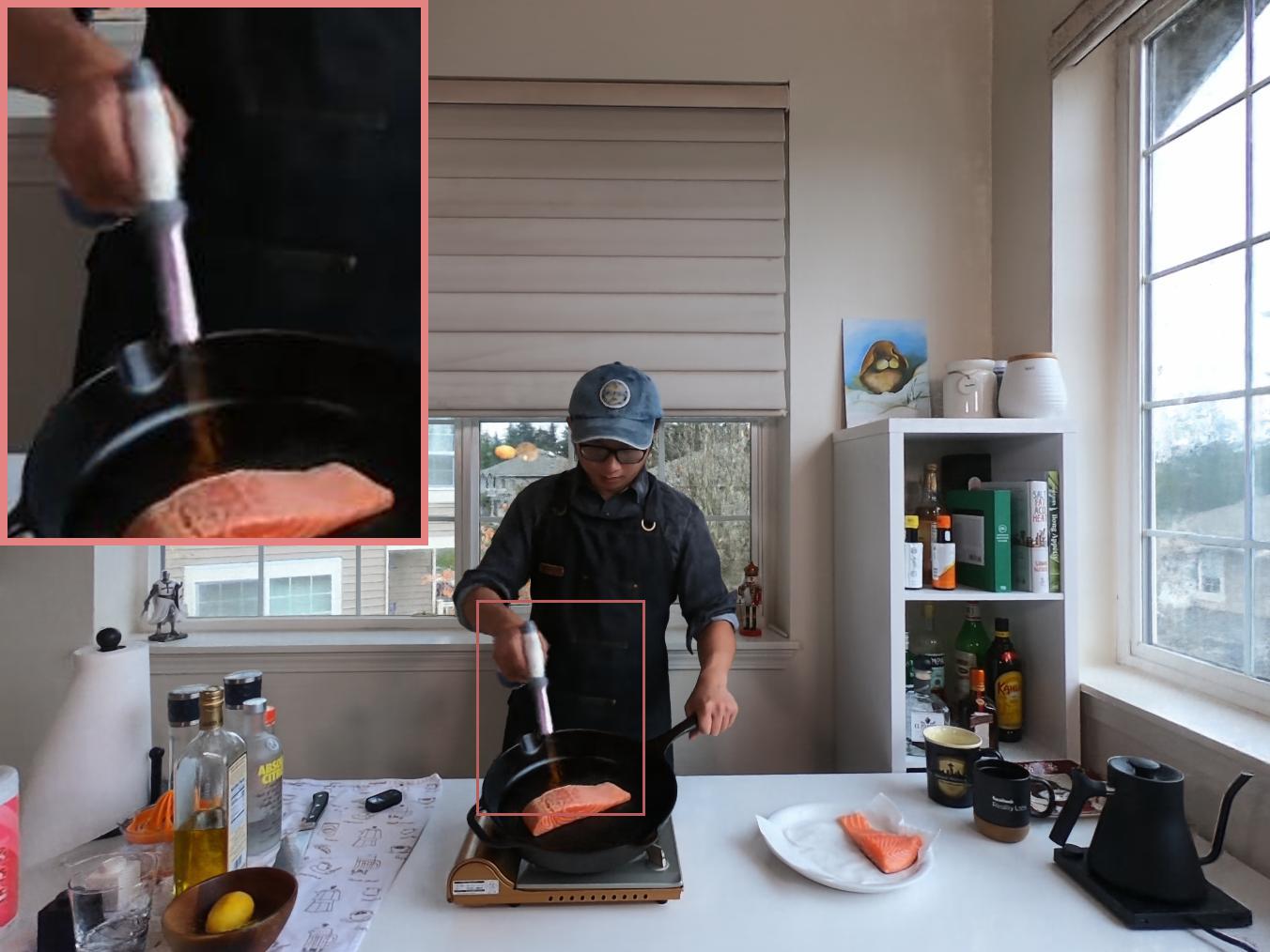}\\DyNeRF\cite{Li_2022_CVPR}\end{tabular} \\
\hspace{-1em}\begin{tabular}{c}\includegraphics[width=0.23\textwidth]{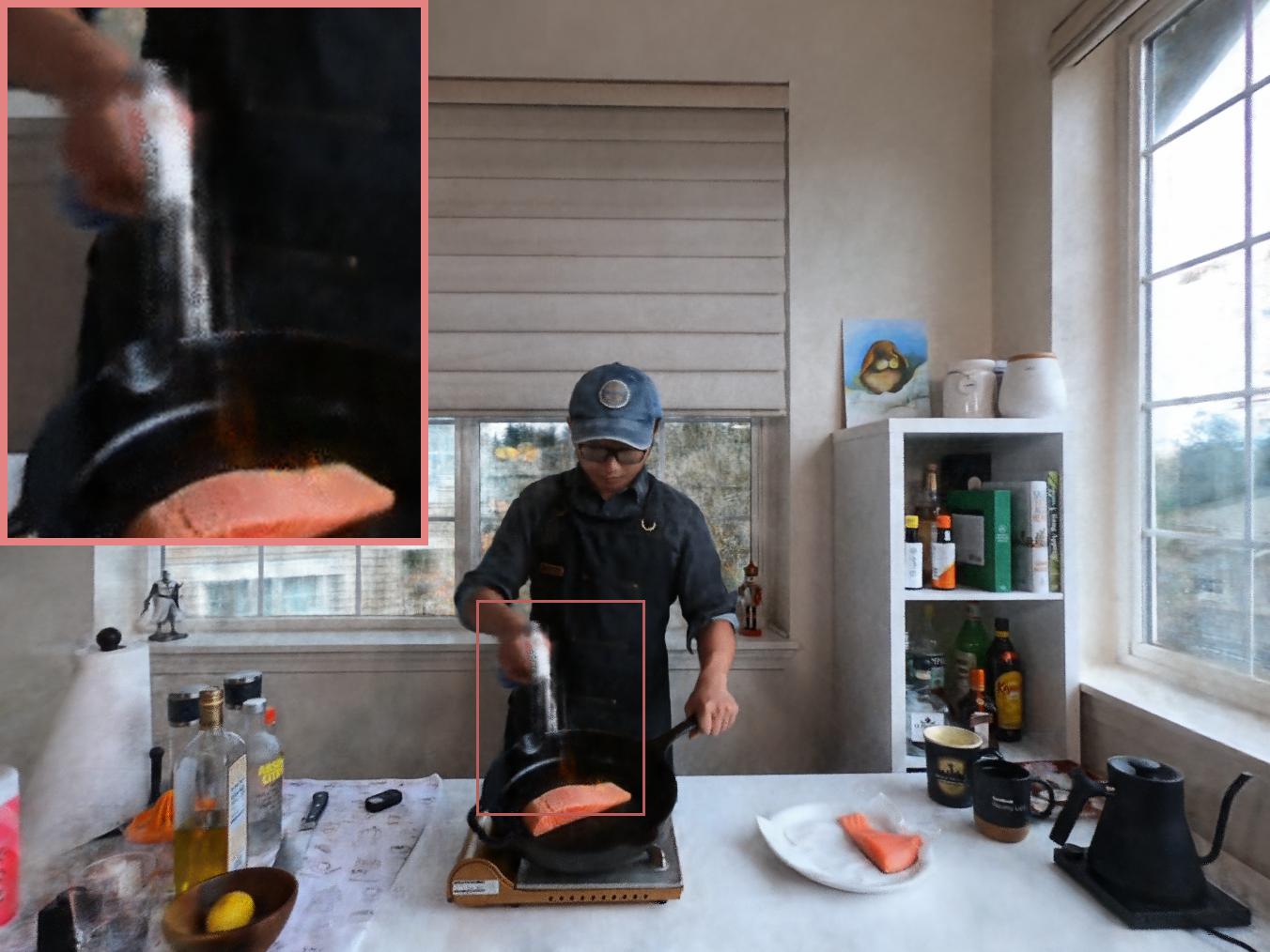}\\Ours-InstantNGP\end{tabular} \hspace{-1em} & 
\hspace{-1em}\begin{tabular}{c}\includegraphics[width=0.23\textwidth]{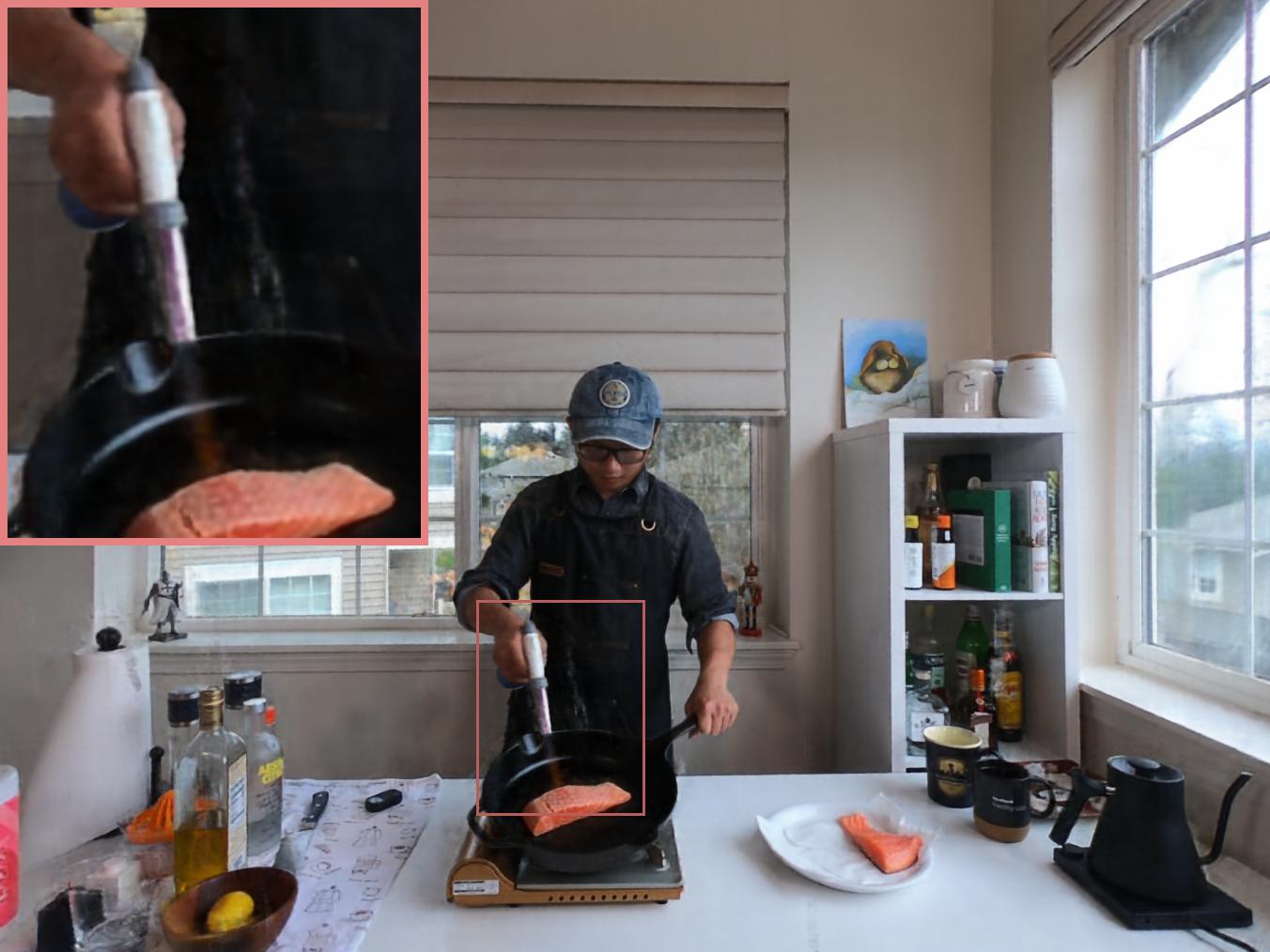}\\Ours-TensoRF-VM\end{tabular}  \\
\end{tabular}
\vspace{-1em}
\caption{Qualitative comparisons on the Plenoptic Video (multi-camera setting) dataset.}\label{fig:n3dv-compare}
\end{figure}

\begin{figure}
\centering
\resizebox*{\linewidth}{!}{
\begin{tabular}{cc}
\includegraphics[height=15em]{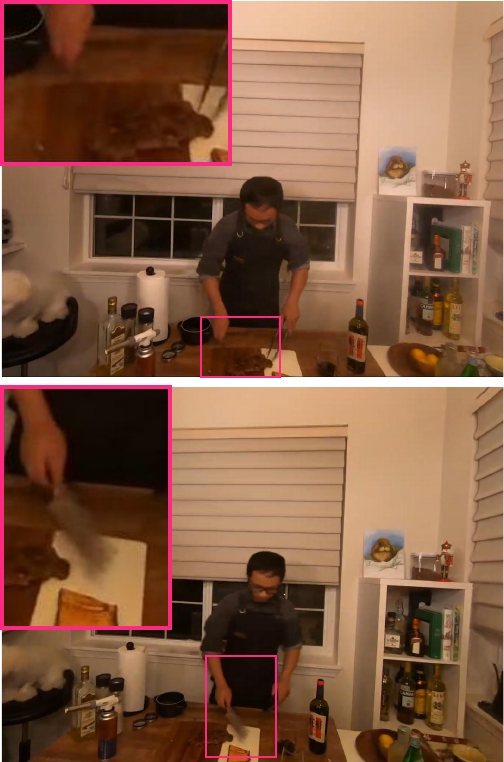}\hspace{-1em} & 
\includegraphics[height=15em]{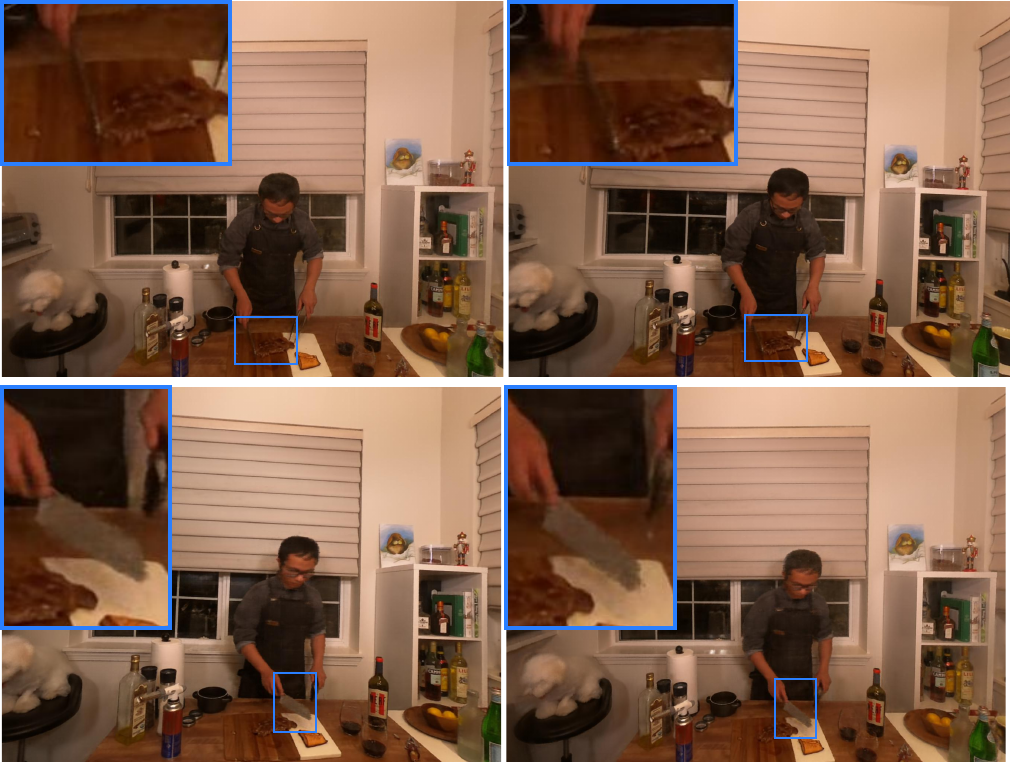} \\
DyNeRF \cite{Li_2022_CVPR} & Ours-TensoRF-VM
\end{tabular}
}
\caption{Comparisons of rendering performance on fast moving objects.}\label{fig:n3dv-compare-fast}
\end{figure}

\paragraph{Datasets.} 
Our method requires only RGB observations of the dynamic scene for reconstruction. Unlike most existing methods, our framework does not require special capturing settings or prior knowledge, and detailed comparisons of the framework's requirements against competitive methods are attached in the supplementary. 
Two multi-camera datasets and one single-camera dataset are used:
\begin{itemize}
  \item \textbf{Immersive Video} \cite{broxton2020immersive} includes synchronized videos from 46 4K fisheye cameras. For the raw video data provided by the authors, each camera has different imaging parameters like exposure and white balance. We select 7 dynamic scenes with relatively similar imaging parameters. We downsample the images to $1280\times960$ in our experiments. The camera with ID 0 (the central camera) is used for validation and the other cameras are used for training.
  \item \textbf{Plenoptic Video} \cite{Li_2022_CVPR} is captured with 21 cameras at a resolution of $2704\times2028$. Different from Immersive Video which mostly focuses on outdoor scenes, Plenoptic Video consists of indoor activities in various lighting conditions. We downsample images to $1352\times1014$ in our experiments. We follow the training and validation camera split provided by \cite{Li_2022_CVPR}. Six scenes are publically available.
  \item \textbf{HyperNeRF} \cite{nerfies,park2021hypernerf} provides only one view for each timestamp in a dynamic scene. The dataset is challenging due to the single-camera setting. We adopt the same training and validation settings as in \cite{park2021hypernerf}: images of $960\times540$ are used for quantitative evaluation and images of $1920\times1080$ are used for qualitative comparisons. There are two capturing settings in HyperNeRF: ``\textit{vrig}'' captures the scene with stereo cameras and training with one camera and validating with the other; ``\textit{interp}'' is a monocular video from a moving camera capturing dynamic scenes.
\end{itemize}
\paragraph{Implementation details.} 
Our framework, as demonstrated by \cref{fig:framework}, is implemented with PyTorch \cite{paszke2019pytorch}. As highlighted in \cref{sec:framework}, our framework is general and any hybrid representation adopting explicit features can be used. We implement our framework with two backbones: InstantNGP \cite{instantngp} and TensoRF \cite{tensorf}. In both of the implementations, the deformation network is a 4-layer MLP with a width of 256. The stationary field $s$ uses a 2-layer MLP with a width of 64. The radiance field $r$ is a 4-layer MLP with a width of 64 and has the same structure as the decoder in the backbone. For InstantNGP based model, the number of levels is 8 and the number of feature dimensions per entry is 4. TensoRF based model follows the same setting as in their experiments on the real forward-facing datasets (\ie, LLFF \cite{llff}). 
For both of the two backbones, we set the number of channels for streaming $k$ to be 1, and loss hyper-parameters $\lambda=0.1$, $\alpha=0.01$.
We follow the default optimization schedule and settings as in the static-scene targeted backbone methods. 
For the two multi-camera datasets, we observe that their frame rates are high and simply modeling every dynamic area as new areas already lead to good temporal interpolation performance, so the deformation is not used by default for efficiency. An ablation is presented for studying the impact of video FPS when modeling with and without deformation decomposition.
Due to the limitation of model sizes, we split a long video into 90-frame clips and trained on these clips separately.
PSNR and SSIM \cite{ssim} are reported for evaluating the rendering performance. 

\begin{table}
\centering
\caption{Quantitative comparisons on Plenoptic Video \cite{Li_2022_CVPR} for multi-camera dynamic scenes.}
\label{tab:n3dv}
\resizebox{\columnwidth}{!}{
\begin{tabular}{lccc}
\toprule 
\multirow{2}*{{Method}}
& \multirow{2}*{{PSNR}$\uparrow$}
& {Training Time}
& {Rendering Time}
\\ 
&& {\small{{(GPU Hours)}}}
& {\small{{(s/img)}}} \\
\midrule
Neural Volumes \cite{lombardi2019neural} & 22.797 & - & - \\
LLFF \cite{llff}  & 23.238 & - & - \\
NeRF-T \cite{Li_2022_CVPR} & 28.448 & - & 90 \\
DyNeRF \cite{Li_2022_CVPR} & 29.580 & 1344 & 90 \\
\midrule
{Ours}-InstantNGP & {30.293} & {5.5} & {10.8} \\
{Ours}-TensoRF-VM & {30.692} & {6} & {22.1} \\
\bottomrule
\end{tabular}
}
\end{table}

\begin{table*}
\centering
\caption{Per-scene quantitative comparisons on HyperNeRF-\textit{vrig} \cite{park2021hypernerf} for single-camera dynamic scenes.}
\label{tab:hypernerf}
\resizebox{\textwidth}{!}{
\begin{tabular}{l|c|cccccccc|cc}
\toprule 
\multirow{2}*{\textbf{Method}} & Rendering Time
& \multicolumn{2}{c}{Broom} & \multicolumn{2}{c}{3D Printer} & \multicolumn{2}{c}{Chicken} & \multicolumn{2}{c}{Peel Banana}  & \multicolumn{2}{|c}{\textbf{Mean}}\\ 
& (s/img)
& PSNR$\uparrow$  & SSIM$\uparrow$ 
& PSNR$\uparrow$  & SSIM$\uparrow$ 
& PSNR$\uparrow$  & SSIM$\uparrow$ 
& PSNR$\uparrow$  & SSIM$\uparrow$ 
& PSNR$\uparrow$  & SSIM$\uparrow$ \\
\midrule
    NeRF~\cite{mildenhall2020nerf} 
    & $\sim$75
    & 19.9&0.653 
    & 20.7&0.780 
    & 19.9&0.777
    & 20.0&0.769 
    & 20.1&0.745 \\
    NV~\cite{lombardi2019neural} 
    & $<$0.03
    & 17.7&0.623 
    & 16.2& 0.665
    & 17.6& 0.615 
    & 15.9&0.380   
    & 16.9&0.571 \\
    NSFF~\cite{nsff}$^*$
    & $\sim$90
    &26.1 &0.871 
    & 27.7& 0.947
    & 26.9& 0.944
    & 24.6& 0.902  
    & 26.3& 0.916 \\
    Nerfies~\cite{nerfies} 
    & $\sim$90
    &19.2 &0.567 
    & 20.6&0.830 
    & 26.7& 0.943 
    & 22.4& 0.872 
    & 22.2& 0.803\\
    HyperNeRF~\cite{park2021hypernerf} 
    & $\sim$90
    &19.3 &0.591 
    & 20.0& 0.821
    & 26.9& 0.948
    & 23.3& 0.896  
    & 22.4& 0.814\\
    \midrule
    Ours-InstantNGP
    & 4.8
    & 21.7 & 0.635
    & 22.9 & 0.810
    & 26.3 & 0.905
    & 24.0 & 0.863
    & 23.7 & 0.803
    \\
\bottomrule
\end{tabular}
}
\end{table*}

\begin{figure*}
\centering
\resizebox{\textwidth}{!}{
\begin{tabular}{ccccccc}
\includegraphics[height=7em]{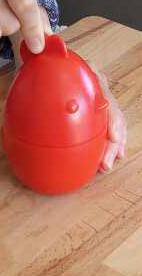} &
\includegraphics[height=7em]{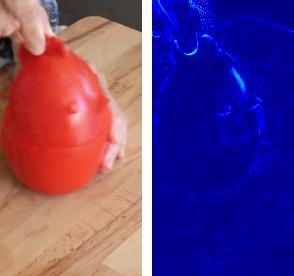} &
\includegraphics[height=7em]{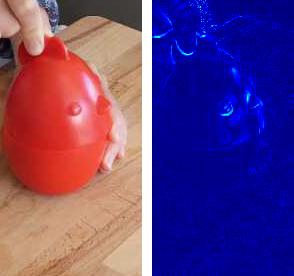} &
\includegraphics[height=7em]{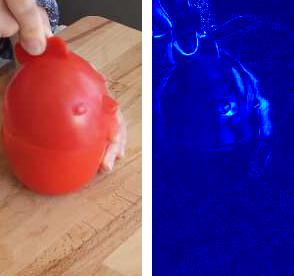} &
\includegraphics[height=7em]{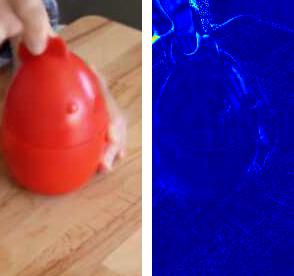} &
\includegraphics[height=7em]{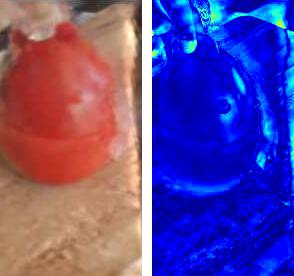} &
\includegraphics[height=7em]{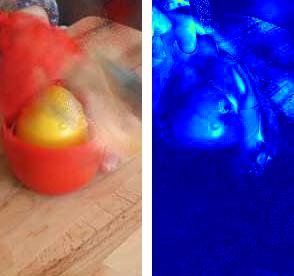} \\
\includegraphics[height=7em]{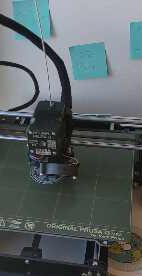} &
\includegraphics[height=7em]{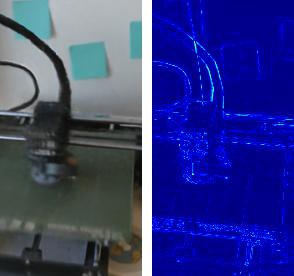} &
\includegraphics[height=7em]{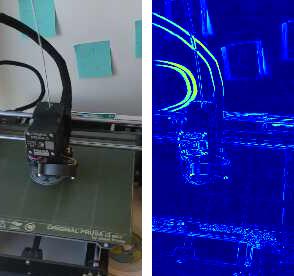} &
\includegraphics[height=7em]{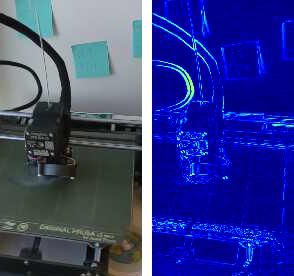} &
\includegraphics[height=7em]{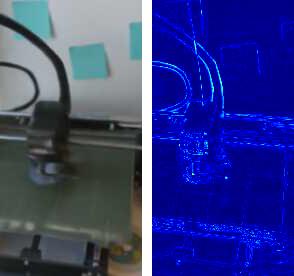} &
\includegraphics[height=7em]{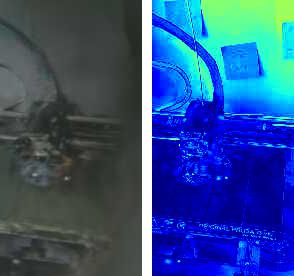} &
\includegraphics[height=7em]{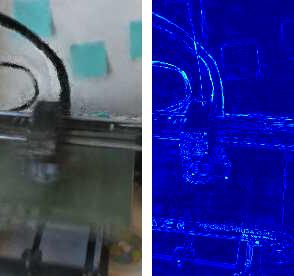} \\
GT & Ours-InstantNGP & HyperNeRF \cite{park2021hypernerf} & Nerfies \cite{nerfies} & NSFF \cite{nsff} & NV \cite{lombardi2019neural} & NeRF \cite{mildenhall2020nerf}
\end{tabular}
}
\vspace{-1em}
\caption{Qualitative comparisons on the HyperNeRF-\textit{vrig} dataset. Error map is demonstrated beside each rendered image. 
}\label{fig:visual-hypernerf}
\end{figure*}

\subsection{Comparison with State-of-The-Art Methods}
DyNeRF \cite{Li_2022_CVPR} and HyperNeRF \cite{park2021hypernerf} are considered for multi- and single-camera settings. Besides the two methods, we also quote the results of other baseline methods reported in their paper. 
\subsubsection{On multi-camera dataset}
In \cref{tab:n3dv}, we report out results with both InstantNGP and TensoRF backbones. Training time and rendering time of DyNeRF are quoted from their paper. Our method reaches a higher PSNR while significantly reducing the training and rendering time. 
We further compare the rendered images in \cref{fig:n3dv-compare}. Images of comparison methods are again quoted from the result images in DyNeRF's paper.  Our method with InstantNGP renders images with 12\% of the time required by DyNeRF while being comparable. Besides, our method with TensoRF-VM achieves better performance on fast-moving objects. As demonstrated in \cref{fig:n3dv-compare-fast}, we compare our rendered results with extracted frames from DyNeRF's result video. Since the code and rendering parameters of DyNeRF are not publically available, we manually select similar camera poses and timestamp for comparison. We can observe that the knife in DyNeRF's results is blurry while our method yields clearer results.

\subsubsection{On single-camera dataset}
A challenging and practical appealing setting is reconstructing and rendering without  per-frame multi-view observations, \ie, capturing with a single camera. 
Our method with TensoRF backbone is not reported on this dataset since we find that the GPU memory required for training is too large with the default model setting. 
We compare our method with SoTA single-camera reconstruction methods in \cref{tab:hypernerf}. Note that NSFF requires data-driven depth and optical flow priors. We can observe that our method outperforms HyperNeRF in terms of PSNR but is slightly worse than HyperNeRF on SSIM. We presume the reason is that our method generates more accurate but less sharp images compared to HyperNeRF. Visual comparisons can be found at \cref{fig:visual-hypernerf}. We can observe that HyperNeRF sometimes has misalignment between the rendered and real images regarding moving objects, such as the wire in the second row. We attribute the misalignment problem to not correctly modeling the deformation. The incorrect modeling is partially caused by treating all pointing as deforming in their representation. Lacking decomposing static and dynamic areas also leads to a flickering background (demonstrated in our video). 
In our method, by decomposing static and dynamic areas, the deformation field is regularized to only model dynamic areas.

\subsection{Ablation Studies}
\subsubsection{Impact of Decomposition}
\begin{figure*}
\centering
\resizebox{\textwidth}{!}{
\begin{tabular}{c}
\includegraphics[width=\textwidth]{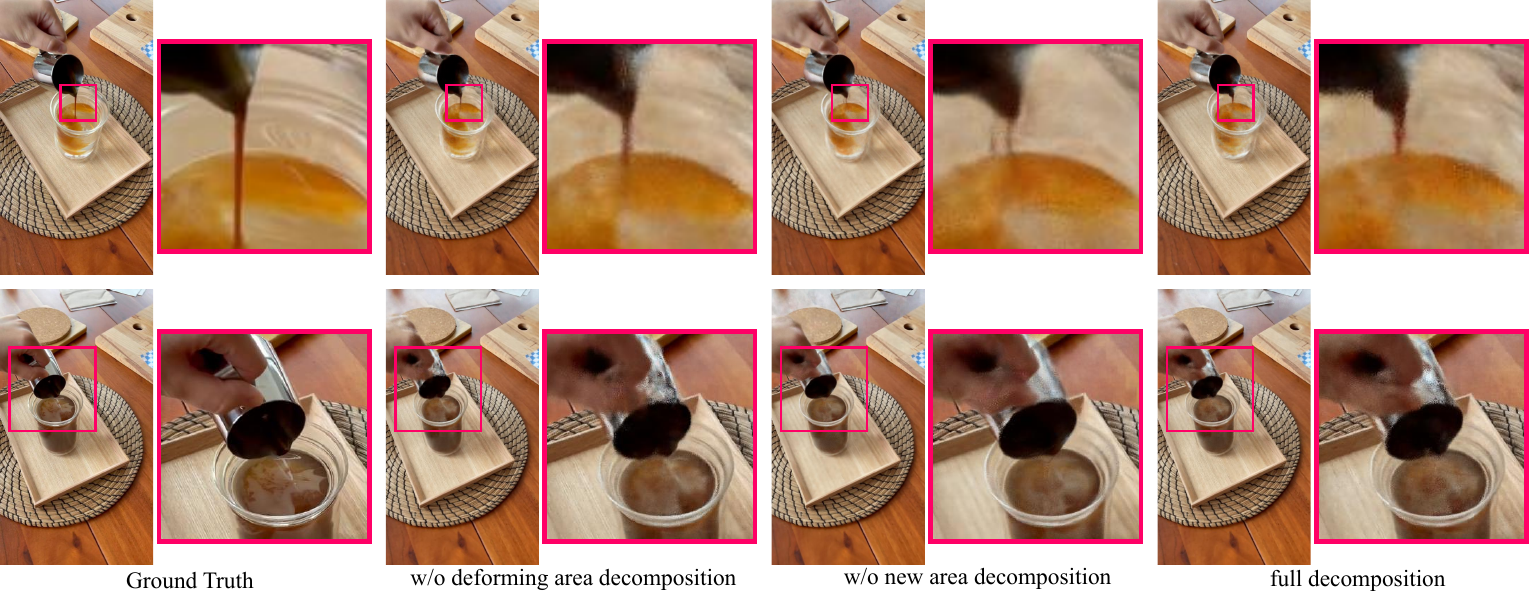}\\
(a) Results after removing deforming and new area decomposition on the single-camera dataset (HyperNeRF).  \\ \\
\includegraphics[width=\textwidth]{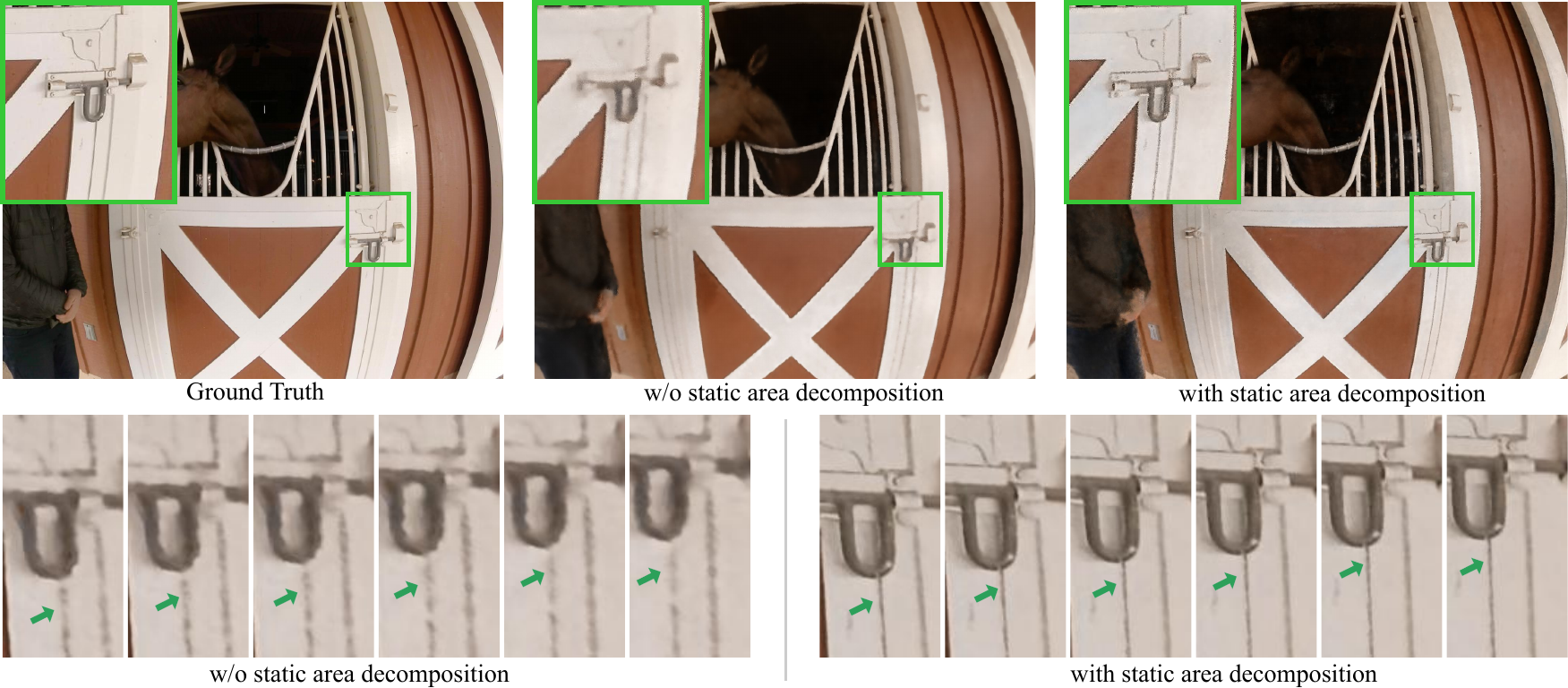}\\
(b) Ablation of static area decomposition on the multi-camera dataset (Immersive Video).  Second row: novel time and view rendering. \\ \\
\includegraphics[width=\textwidth]{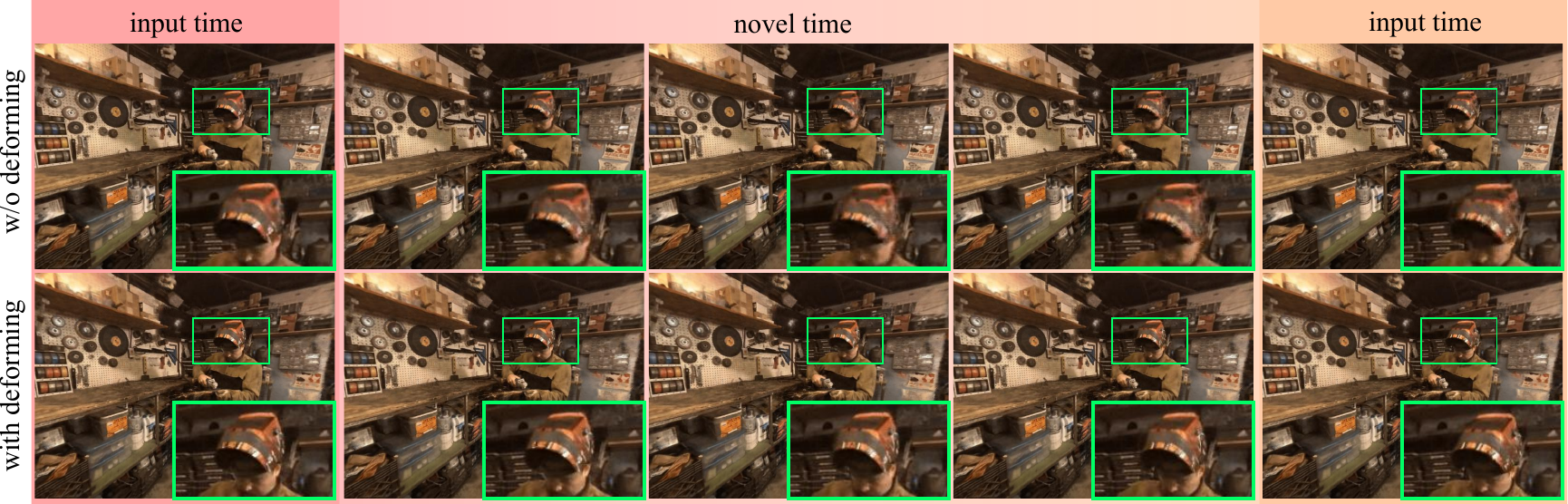}\\
(c) Ablation of deforming area decomposition on the multi-camera dataset (Immersive Video). Every 8 frames are used for training.
\end{tabular}
}
\caption{Ablation of scene decompositions with InstantNGP as the backbone. (a) For the single-camera dataset (HyperNeRF), the full decomposition can well reconstruct the newly generated fluid and moving cup. (b) For multi-camera dataset (Immersive Video), static area decomposition leads to a clearer background and suppress flickering background. (c) When the recording frame rate is low ($\sim$4 FPS) and objects move faster, deformation decomposition can help generate smoother temporal interpolation results.}\label{fig:ablation}
\end{figure*}

We first study the necessity of the proposed three categories for decomposition. 
Visual comparisons of different decomposition variants are demonstrated in \cref{fig:ablation} and quantitative results are reported in \cref{tab:ablation}. First, we study the impact of decomposing deforming and new areas on a single-camera dataset. We can observe from \cref{fig:ablation}(a) that removing new area decomposition leads to failure of modeling the newly poured out espresso and removing deforming area decomposition leads to a blurred hand and cup.
Second, we study the impact of our decomposition on the multi-camera dataset. \cref{fig:ablation}(b) demonstrates that the static area becomes sharper after decomposing static areas. Besides, without static area decomposition, we observe that the background is flickering as we render images with novel time and view.

Finally, we study the impact of deforming area decomposition on the multi-camera dataset. This ablation study is motivated by the observation that rendering with and without deformation modeling leads to little difference (PSNR difference less than 0.1). We presume that this is because the motion of objects between frames is small from cameras with a high FPS recording rate. Therefore modeling all dynamic areas with a newness field can still produce a smooth interpolation. We manually downsample the frame sampling rate for training in \cref{fig:ablation}(c) to enlarge the motion between frames. We can observe that without deformation modeling the moving helmet becomes first disappeared and then reappeared when interpolating between two training timestamps. As a comparison, the content of the helmet is well preserved if the deformation is modeled.

\begin{figure}[b]
\centering
\vspace{-2.5em}
\includegraphics[width=\columnwidth,trim={7em 0em 12.8em 2em},clip]{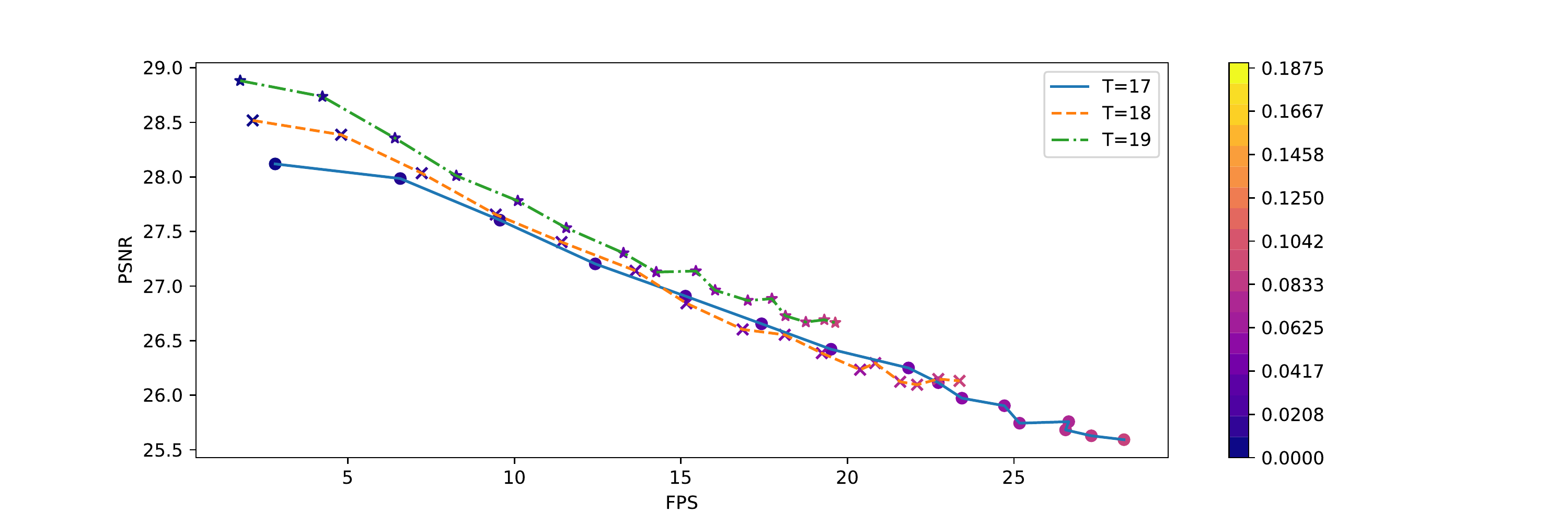}
\vspace{-2.5em}
\caption{Rendering speed and quality tradeoff with InstantNGP backbone. The color of the marker demonstrates the value of exponential stepping during ray marching.}\label{fig:realtime}
\end{figure}

\begin{figure}[t]
\centering
\includegraphics[width=\columnwidth]{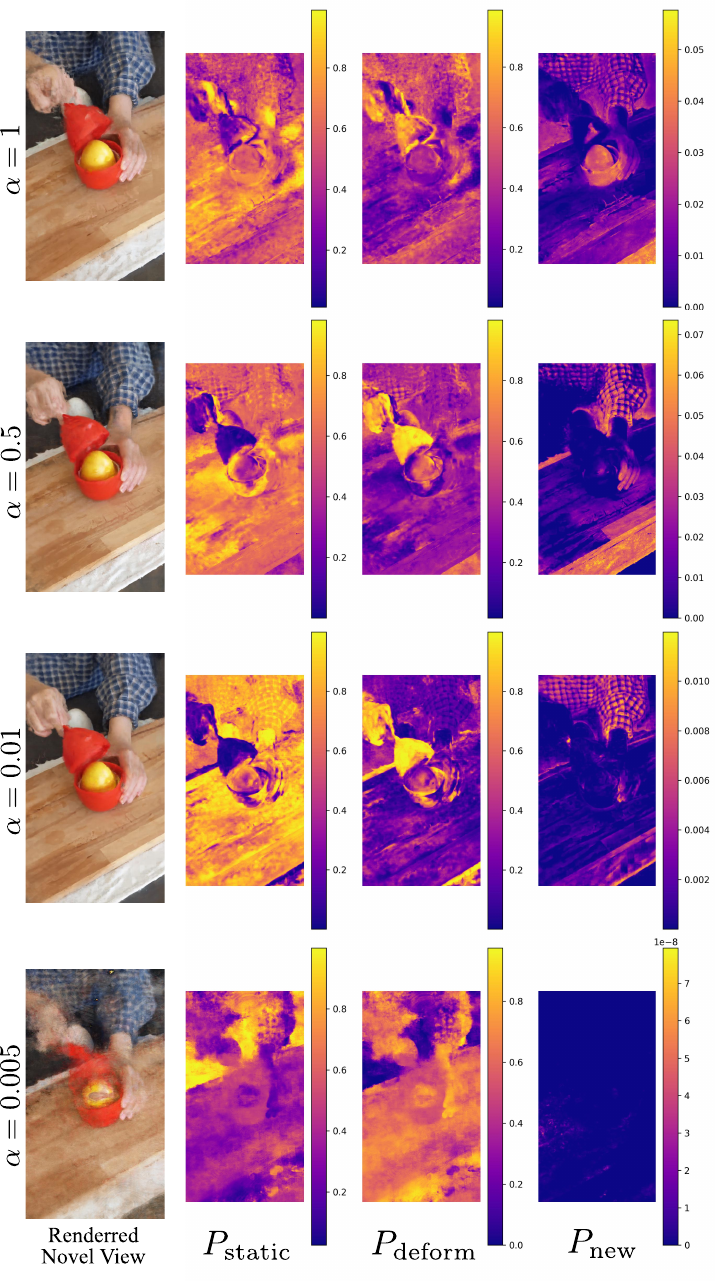}
\caption{Visualization of the decomposition and rendering results on HyperNeRF under different scene regularization settings.}\label{fig:prior}
\end{figure}

\begin{figure}[t]
\centering
\includegraphics[width=\columnwidth]{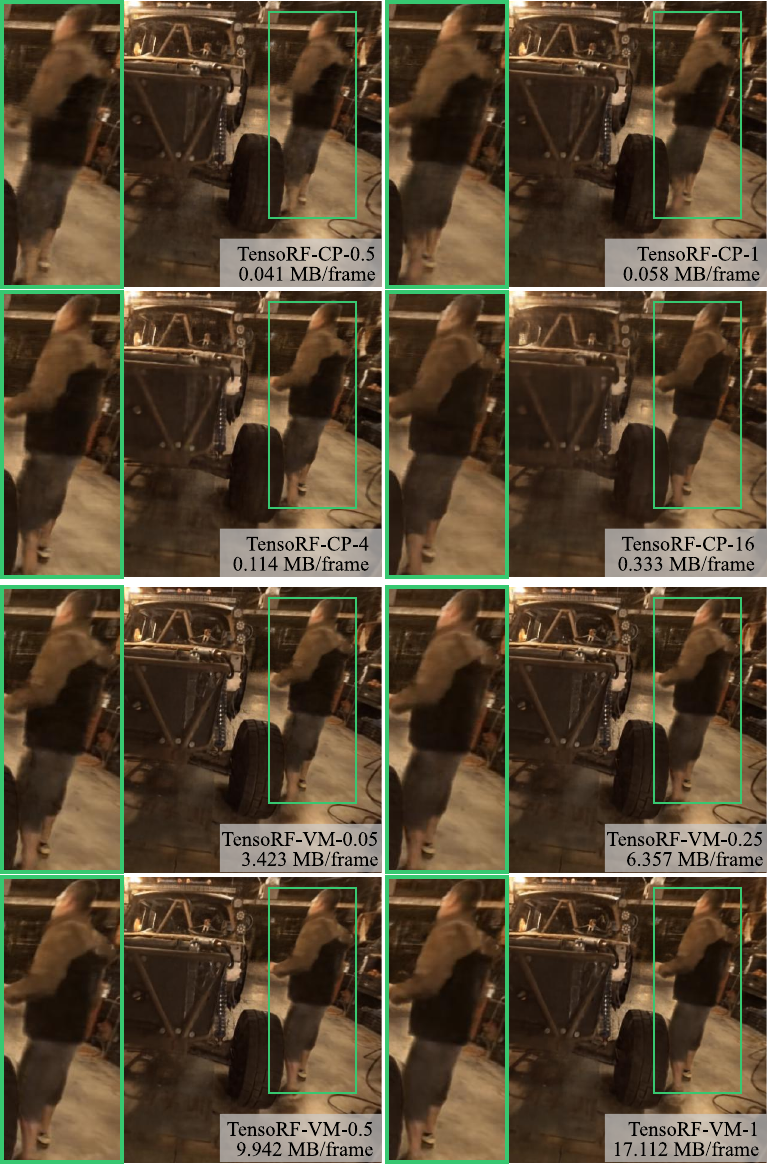}
\caption{Comparison of results with different bitrate budgets for streaming. On each image, we report the model name followed by the value of $k$ and the bitrate respectively.}\label{fig:bitrate}
\end{figure}

\begin{table}
\centering
\caption{Quantitative results of different decomposition variants with InstantNGP as the backbone.}\label{tab:ablation}
\resizebox*{\columnwidth}{!}{
\begin{tabular}{lcccc}
\toprule
Method & PSNR$\uparrow$ & SSIM$\uparrow$ & LPIPS$_{\mathrm{VGG}}\downarrow$  & LPIPS$_{\mathrm{Alex}}\downarrow$ \\
\midrule
\multicolumn{4}{l}{\textcolor{gray}{\textit{on HyperNeRF-interp}}}\\
w/o deforming & 28.2 & 0.820 & 0.358 & 0.220 \\
w/o new & 28.2 & 0.837 & 0.323 & 0.188 \\
full & 29.2 & 0.858 & 0.294 & 0.163 \\
\cmidrule(lr){1-5}
\multicolumn{4}{l}{\textcolor{gray}{\textit{on Immersive Video (``Horse'')}}}\\
w/o static & 27.0 & 0.860 & 0.423 & 0.255 \\
w/ static & 27.4 & 0.871 & 0.425 & 0.295 \\\cmidrule(lr){1-5}
\multicolumn{4}{l}{\textcolor{gray}{\textit{on Immersive Video (``Welder'' with every 8 frames)}}}\\
w/o deforming &  26.1 & 0.826 & 0.366 & 0.195 \\
w/ deforming & 25.2 & 0.800 & 0.327 & 0.168 \\
\bottomrule
\end{tabular}
}
\end{table}

\subsubsection{Scene Decomposition Regularizing}
In our method, we use $\alpha$ to balance the ratio of being deforming and new in \cref{eq:lossreg}. A larger $\alpha$ encourages the scene to contain fewer deforming areas. 
As introduced in the previous section, single-camera datasets are more sensitive to the deformation field, thus we study the impact of $\alpha$ in \cref{fig:prior} on a scene from HyperNeRF.
We can observe that over-suppressing deforming areas ($\alpha=1$) lead to blurry moving objects and under-suppressing deforming areas ($\alpha=0.005$) leads to a noisy scene. The reason behind the blur from large $\alpha$ is the same as the second row in \cref{fig:toy-example} and \cref{fig:ablation}(c): falsely modeling a moving object as first-disappear-then-reappear. A good practice is that we can start with a relatively large $\alpha$ to penalize deforming areas and then gradually allow areas to deform by decreasing $\alpha$. 

\subsubsection{Streaming Bitrates}
An important metric for a streaming service is the bitrate. To render a new frame, the user is usually sensitive to the new data needed to download. We can easily tune the bitrate requirements in our method by setting the value of $k$. 
In \cref{tab:bitrate}, we report the bitrate for streaming a new frame with different $k$ values. The testing data is a sequence from Immersive Video with 90 frames. Note that $k$ denotes new channels needed for rendering a new frame and rendering the first frame still follows the channels required for static scenes (96 for TensoRF-CP and 4 for TensoRF-VM). For fair comparisons, bitrate is computed by the total model size over the number of frames. 

The TensoRF-CP based model achieves low bitrate and reasonable performance, while the cost of TensoRF-VM is higher but the performance gain is obvious. We further present rendering results in \cref{fig:bitrate}. We can observe clearer details of the background (\ie, car) and the moving objects (\ie, person) with increased bitrate budgets. The above results validate the extensibility of our framework.

\begin{table}
\centering
\caption{Quantitative results of rendering with different bitrate budgets for streaming.}\label{tab:bitrate}
\resizebox*{\columnwidth}{!}{
\begin{tabular}{lccccc}
\toprule
Method & PSNR$\uparrow$ & SSIM$\uparrow$ & LPIPS$_{\mathrm{VGG}}\downarrow$ & \begin{tabular}{c}Bitrates\\(MB/frame)\end{tabular} \\
\midrule
Immersive \cite{broxton2020immersive} & - & - & - & $\sim$0.5 \\
\cmidrule(lr){1-5}
\multicolumn{4}{l}{\textcolor{black}{\textit{Ours-TensoRF-CP}}}\\
k=0.50 & 25.200 & 0.754 & 0.284 & 0.041 \\
k=1.00 & 25.798 & 0.846 & 0.264 & 0.058 \\
k=4.00 & 25.870 & 0.835 & 0.266 & 0.114 \\
k=16.00 & 25.885 & 0.857 & 0.244 & 0.333 \\
\cmidrule(lr){1-5}
\multicolumn{4}{l}{\textcolor{black}{\textit{Ours-TensoRF-VM}}}\\
k=0.05 & 26.093 & 0.866 & 0.184 & 3.423 \\
k=0.25 & 26.032 & 0.865 & 0.188 & 6.357 \\
k=0.50 & 26.187 & 0.872 & 0.192 & 9.942 \\
k=1.00 & 26.203 & 0.878 & 0.173 & 17.112 \\
\bottomrule
\end{tabular}
}
\end{table}

\subsubsection{Rendering Speed and Quality}
The performance of our framework is highly correlated with the chosen backbones. Thus, in our method, there exists a tradeoff between rendering speed and quality, mainly affected by predefined model size and rendering hyper-parameters. In \cref{fig:realtime}, we present the rendering FPS and PSNR with different hyper-parameter settings. We consider two parameters: $T$ for the hash table size and the stepping value during ray marching. Scenes from the Immersive Video dataset are considered. We can observe that our method inherits the flexibility of the backbone and we can easily tune the parameters to obtain the desired speed and quality.

\begin{figure}[h!]
\resizebox{\columnwidth}{!}{
\begin{tabular}{ccc}
\includegraphics[width=0.33\columnwidth]{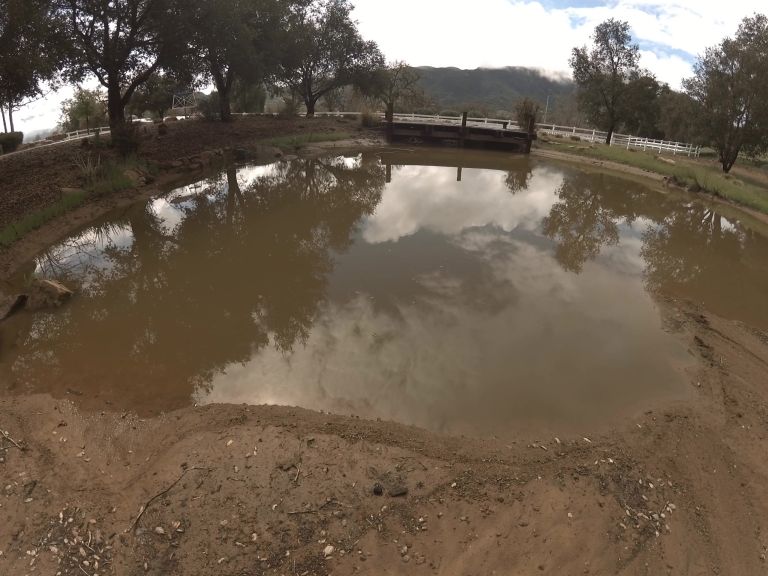}&\hspace*{-1.5em}
\includegraphics[width=0.33\columnwidth]{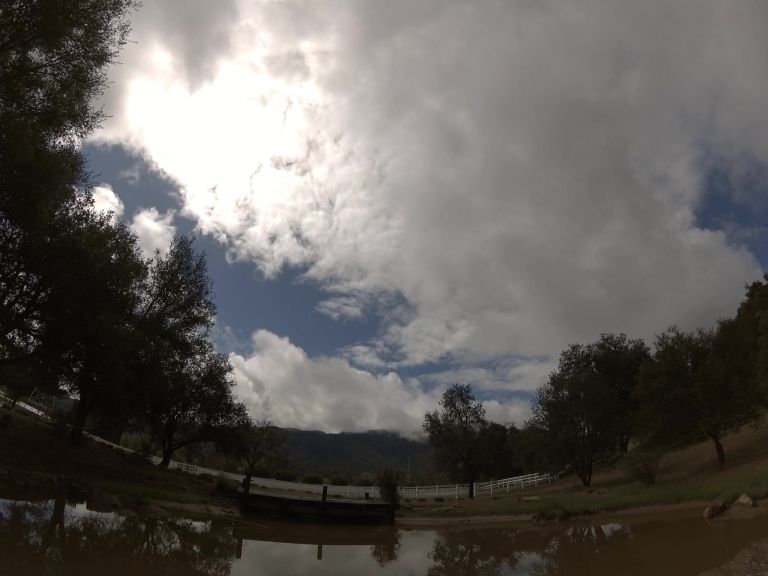}&\hspace*{-1em}
\includegraphics[width=0.33\columnwidth]{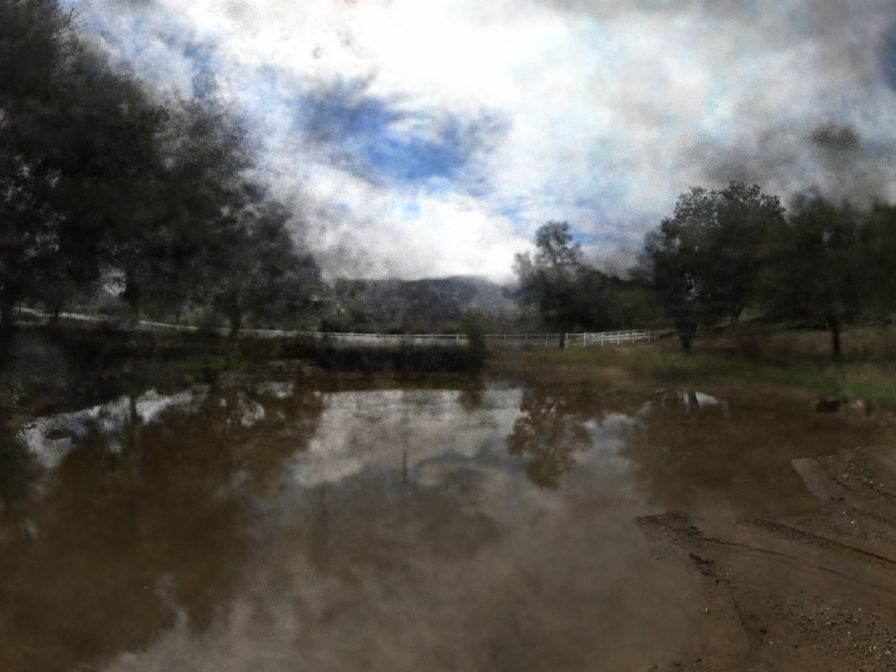}\\
\multicolumn{2}{c}{Input images} & Novel view
\end{tabular}
}
\caption{Our method fails when inputs are with different imaging configurations (\eg, exposure).}
\label{fig:fail}
\end{figure}

\section{Limitation and Failure Cases}
Our method models each frame in the scene with local feature channels, which enables streaming but limits the representation of long-range repeated activities. For example, the activity of pouring espresso in \cref{fig:points} may repeat several times in a scene. Further modeling the repeating activities can reduce redundancy and improve the reconstruction quality by leveraging all the views of the same object. Moreover, our method assumes input multi-view images are with the same camera imaging configuration (\eg, exposure). A failure example from Immersive Video is demonstrated in \cref{fig:fail}. Though view dependency can still be modeled in the framework, the model tends to generate floating points to overfit the training views. Recent progress that considers the photography process \cite{martin2021nerf,mildenhall2021rawnerf} may help solve the issue.

\section{Conclusion}
We present a framework for representing dynamic scenes from both multi- and single-camera captured images. The key components of our framework are the decomposition module and the feature streaming module. The decomposition module decomposes the scene into static, deforming, and new areas. A sliding window based hybrid representation is then designed for efficiently modeling the decomposed neural fields. 
Experiments on multi- and single-camera datasets validate our method's efficiency and effectiveness. Extensive ablation studies further provide insight into the model design, such as the necessity of modeling deformation in large-motion scenes captured by camera arrays.

{\small
\bibliographystyle{ieee_fullname}
\bibliography{nerf}
}

\onecolumn
\newpage
\section*{Appendix}
Per-scene performance evaluations are presented in \cref{tab:immersive-perscene} (for Immersive Video) and \cref{tab:n3dv-perscene} (for Plenoptic Video).
\begin{table*}[h]
\small
\centering
\begin{tabular}{lcccc}
\toprule
Scene & PSNR & SSIM & LPIPS$_{\mathrm{Alex}}$  & LPIPS$_{\mathrm{VGG}}$ \\
\midrule
\rowcolor{Gray}
\textsc{01\_Welder} & 25.568 & 0.818 & 0.289 & 0.420 \\
\rowcolor{Gray}
\textsc{02\_Flames} & 26.554 & 0.842 & 0.154 & 0.271 \\
\textsc{03\_Dog} & 18.764 & 0.579 & 0.463 & 0.515 \\
\rowcolor{Gray}
\textsc{04\_Truck} & 27.021 & 0.877 & 0.164 & 0.311 \\
\rowcolor{Gray}
\textsc{05\_Horse} & 27.416 & 0.871 & 0.295 & 0.425 \\
\textsc{06\_Goats} & 23.023 & 0.794 & 0.278 & 0.386 \\
\textsc{07\_Car} & 19.656 & 0.624 & 0.393 & 0.501 \\
\textsc{08\_Pond} & 18.132 & 0.728 & 0.352 & 0.487 \\
\rowcolor{Gray}
\textsc{09\_Alexa\_Meade\_Exhibit} & 24.549 & 0.869 & 0.151 & 0.278 \\
\rowcolor{Gray}
\textsc{10\_Alexa\_Meade\_Face\_Paint\_1} & 27.772 & 0.916 & 0.147 & 0.314 \\
\rowcolor{Gray}
\textsc{11\_Alexa\_Meade\_Face\_Paint\_2} & 27.352 & 0.902 & 0.152 & 0.326 \\
\textsc{12\_Cave} & 21.825 & 0.715 & 0.314 & 0.381 \\
\textsc{13\_Birds} & 13.776 & 0.826 & 0.318 & 0.451 \\
\textsc{14\_Puppy} & 15.594 & 0.791 & 0.305 & 0.482 \\
\midrule
\textsc{Mean (All)} & 22.643 & 0.796 & 0.269 & 0.396 \\
\rowcolor{Gray}
\textsc{Mean (Selected 7)} & 26.604 & 0.870 & 0.193 & 0.335 \\
\bottomrule
\end{tabular}
\caption{Per scene performance on the Immersive Video dataset. The 7 scenes marked with gray background are the scenes that we find consistent among cameras.}\label{tab:immersive-perscene}
\end{table*}

\begin{table*}[h]
\small
\centering
(a) Ours-InstantNGP\\
\begin{tabular}{lcccc}
\toprule
Scene & PSNR & SSIM & LPIPS$_{\mathrm{Alex}}$  & LPIPS$_{\mathrm{VGG}}$ \\ \midrule
\textsc{coffee martini} & 32.053 & 0.938 & 0.115 & 0.279 \\
\textsc{cook spinach} & 32.064 & 0.930 & 0.116 & 0.284 \\
\textsc{cut roasted beef} & 31.830 & 0.928 & 0.119 & 0.287 \\
\textsc{flame salmon} & 26.140 & 0.849 & 0.233 & 0.379 \\
\textsc{flame steak} & 27.361 & 0.867 & 0.215 & 0.355 \\
\textsc{sear steak} & 32.310 & 0.940 & 0.111 & 0.272 \\
\midrule
\textsc{Mean} & 30.293 & 0.909 & 0.152 & 0.309 \\
\bottomrule
\end{tabular}
\vspace{1em}

(b) Ours-TensoRF-VM\\
\begin{tabular}{lcccc}
\toprule
Scene & PSNR & SSIM & LPIPS$_{\mathrm{Alex}}$  & LPIPS$_{\mathrm{VGG}}$ \\ \midrule
\textsc{coffee martini} & 31.534 & 0.951 & 0.085 & 0.187 \\
\textsc{cook spinach} & 30.557 & 0.929 & 0.113 & 0.226 \\
\textsc{cut roasted beef} & 29.353 & 0.908 & 0.144 & 0.221 \\
\textsc{flame salmon} & 31.646 & 0.940 & 0.098 & 0.203 \\
\textsc{flame steak} & 31.932 & 0.950 & 0.088 & 0.190 \\
\textsc{sear steak} & 29.129 & 0.908 & 0.138 & 0.227 \\
\midrule
\textsc{Mean} & 30.692 & 0.931 & 0.111 & 0.209  \\
\bottomrule
\end{tabular}
\caption{Per scene performance on the Plenoptic Video dataset.}\label{tab:n3dv-perscene}
\end{table*}

\end{document}